\documentclass[10pt,twocolumn,letterpaper]{article}

\usepackage[pagenumbers]{cvpr} 

\usepackage{booktabs}
\definecolor{cvprblue}{rgb}{0.21,0.49,0.74}
\definecolor{plusred}{RGB}{192, 0, 0}
\definecolor{darkblue}{RGB}{60, 90, 200}
\definecolor{lightblue}{RGB}{214, 228, 255}
\usepackage[table]{xcolor}
\usepackage[pagebackref,breaklinks,colorlinks,allcolors=cvprblue]{hyperref}
\usepackage{multirow}
\usepackage{amsmath,amssymb,amsfonts}  
\usepackage{graphicx}  
\usepackage{bbding}
\usepackage{arydshln}
\usepackage{tcolorbox}

\definecolor{kevin}{rgb}{0.74,0.49,0.21}

\definecolor{deepgreen}{RGB}{0,100,0} %

\definecolor{zijin}{RGB}{155,0,0}

\def\paperID{5488} 
\def\confName{CVPR}
\def\confYear{2026}

\title{Generative Visual Chain-of-Thought for Image Editing}

\author{
Zijin Yin\textsuperscript{1,2,3†} \quad
Tiankai Hang\textsuperscript{3} \quad
Yiji Cheng\textsuperscript{3} \quad
Shiyi Zhang\textsuperscript{3} \quad
Runze He\textsuperscript{3} \quad
Yu Xu\textsuperscript{3} \quad \\
Chunyu Wang\textsuperscript{3‡} \quad
Bing Li\textsuperscript{4} \quad
Zheng Chang\textsuperscript{1} \quad
Kongming Liang\textsuperscript{1,2§} \quad
Qinglin Lu\textsuperscript{3} \quad
Zhanyu Ma\textsuperscript{1,2} \quad \\
\textsuperscript{1}Beijing University of Posts and Telecommunications \quad \\
\textsuperscript{2}Beijing Key Laboratory of Multimodal Data Intelligent Perception and Governance \quad \\
\textsuperscript{3}Tencent Hunyuan \quad
\textsuperscript{4}King Abdullah University of Science and Technology \\
\url{https://pris-cv.github.io/GVCoT/}
}

\begin{document}
\twocolumn[{
\renewcommand\twocolumn[1][]{#1}%
\maketitle
\vspace{-2em}
\begin{center}
    \centering
    \captionsetup{type=figure}
    \includegraphics[width=\textwidth]{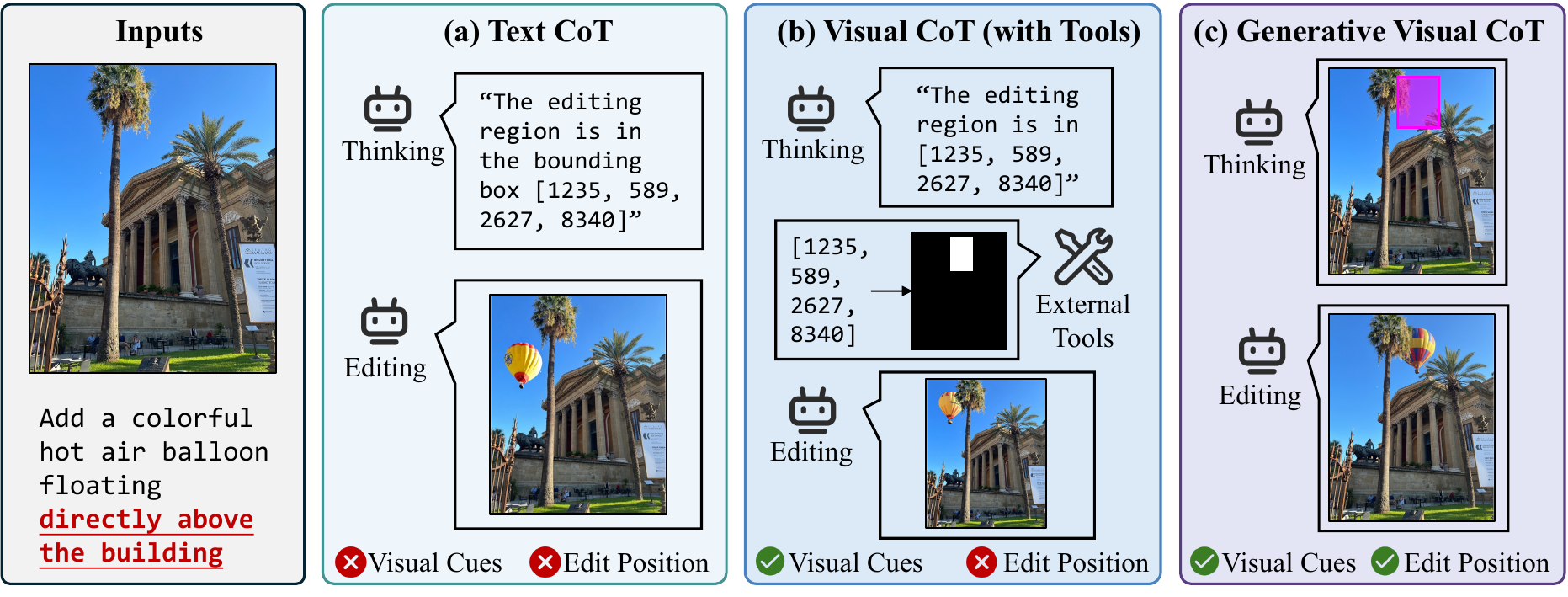}
    \vspace{-1.5em}
    \captionof{figure}{\textbf{Generative Visual Chain-of-Thought (GVCoT).} A comparison of three reasoning paradigms: (a) \textcolor[RGB]{67,149,147}{\textbf{Text CoT}}, which reasons purely within the text space; (b) \textcolor[RGB]{74,131,192}{\textbf{Visual CoT (with Tools)}}, which leverages external tools to highlight target regions; and (c) \textcolor[RGB]{90,60,131}{\textbf{Our GVCoT}}, which performs native visual reasoning via a generative diffusion process within a unified space.}
    \label{fig:teaser}
\end{center}
}]

\renewcommand{\thefootnote}{}
\footnotetext{
\hspace{-1.8em}\textsuperscript{†} Work done during internship at Tencent Hunyuan. \\
\textsuperscript{‡} Project leader. \\
\textsuperscript{§} Corresponding author. liangkongming@bupt.edu.cn
}

\begin{abstract}
Existing image editing methods struggle to perceive where to edit, especially under complex scenes and nuanced spatial instructions.
To address this issue, we propose Generative Visual Chain-of-Thought (GVCoT), a unified framework that performs native visual reasoning by first generating spatial cues to localize the target region and then executing the edit. 
Unlike prior text-only CoT or tool-dependent visual CoT paradigms, GVCoT jointly optimizes visual tokens generated during the reasoning and editing phases in an end-to-end manner. This way fosters the emergence of innate spatial reasoning ability and enables more effective utilization of visual-domain cues.
The main challenge of training GCVoT lies in the scarcity of large-scale editing data with precise edit region annotations; to this end, we construct GVCoT-Edit-Instruct, a dataset of 1.8M high-quality samples spanning 19 tasks.
We adopt a progressive training strategy: supervised fine-tuning to build foundational localization ability in reasoning trace before final editing, followed by reinforcement learning to further improve reasoning and editing quality.
Finally, we introduce SREdit-Bench, a new benchmark designed to comprehensively stress-test models under sophisticated scenes and fine-grained referring expressions. 
Experiments demonstrate that GVCoT consistently outperforms state-of-the-art models on SREdit-Bench and ImgEdit. We hope our GVCoT will inspire future research toward interpretable and precise image editing.
\end{abstract}    
\vspace{-0.5em}
\section{Introduction}
\label{sec:intro}

Recent advances in large-scale datasets and training have enabled significant progress in instruction-guided image editing, through both unified understanding-generation models \cite{deng2025bagel,xie2025show,wang2025ovis,lin2025uniworld} and diffusion-based approaches \cite{wu2025qwenimagetechnicalreport,labs2025flux1kontextflowmatching,liu2025step1x,brooks2023instructpix2pix,promptenhancer}. However, these methods still struggle to localize intended edit regions reliably under complex scenarios, such as tasks involving intricate spatial relations, images with multiple entities, and finely nuanced instructions. 

Several studies \cite{wang2025vgr,guo2025seed1,zhang2025chain} have shown that inference-time scaling, such as Chain-of-Thought (CoT)~\cite{wei2022chain}, improves performance on complex tasks. 
Motivated by this, GoT-R1 \cite{duan2025got,fang2025got} adopts such a strategy into image editing, \textit{i.e.}, predicting target location coordinates within the textual CoT, as illustrated in Fig.~\ref{fig:teaser} (a). 
However, it remains a linguistic proxy and therefore does not fully leverage spatial information within the visual domain. Cognitive science suggests an alternative view: visual reasoning is an inherently modality-specific capacity \cite{Lawrence1999Perceptual}. A skilled artist “paints twice”, first imagining in the mind, then drawing on the canvas. This raises a new question: \textit{Can integrating reasoning through visual intermediates improve image editing more effectively than solely using textual reasoning results?}

\begin{figure}
    \centering
    \includegraphics[width=\linewidth]{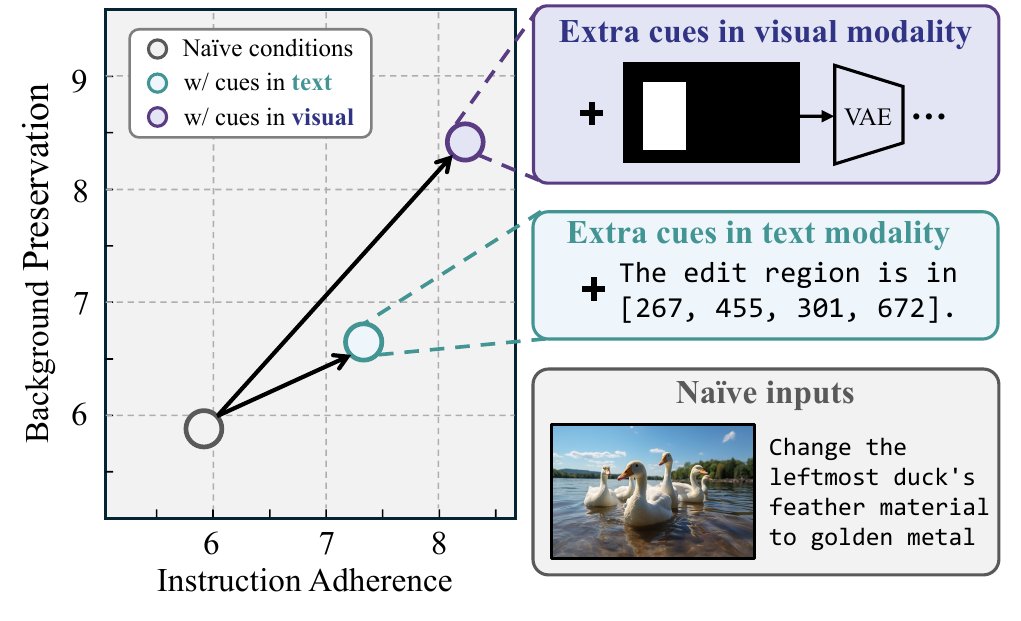}
    \vspace{-2em}
    \caption{ 
    \textbf{Comparing spatial cue representation for image editing on ImgEdit~\cite{ye2025imgedit}.}
    We study two ways of injecting spatial information: 
    (1) \textcolor[RGB]{67,149,147}{\textbf{text modality}} uses bounding box coordinates, and (2) \textcolor[RGB]{90,60,163}{\textbf{visual modality}} providing a binary mask. 
    Providing spatial information in the visual modality yields a greater improvement in both instruction adherence and background preservation.
    }
    \vspace{-1.5em}
    \label{fig:teaser_1}
\end{figure}

To investigate this question, we conduct a preliminary study comparing two methods of providing spatial cues: (1) bounding-box coordinates in text modality, and (2) bounding-box masks in visual modality. As shown in Fig.~\ref{fig:teaser_1}, visual modality cues yield superior instruction adherence and better background preservation. These findings establish that \textit{visual-level spatial cues are more effective than text-level cues for image editing.} 
One straightforward approach to incorporate such cues is through an agentic pipeline that integrates external visual aids (such as cropping, zooming, or tool-generated masks) into reasoning traces \cite{li2025imagine,su2025pixel,zheng2025deepeyes}, as illustrated in Fig.~\ref{fig:teaser} (b).
However, this paradigm is fundamentally limited by the expressiveness of external tools. Since the reasoning remains text-driven, the model cannot develop \textit{innate visual reasoning} capabilities.

In this paper, we propose \textbf{G}enerative \textbf{V}isual \textbf{C}hain-\textbf{o}f-\textbf{T}hought (\textbf{GVCoT}), a novel framework that enables a unified model to \textit{generate visual spatial cues} as intermediate reasoning steps during image editing (see Fig.~\ref{fig:teaser} (c)).
Specifically, the process begins by identifying the editing region by drawing masks onto the input image, which corresponds to the visual thought, followed by the image editing step. The main advantage is that, by directly supervising the visual tokens generated during the reasoning process with a diffusion loss \cite{song2020score,ho2020denoising}, GVCoT integrates reasoning and editing into a unified end-to-end learning framework, thereby facilitating a more stable and effective emergence of intrinsic visual reasoning ability.

The key challenge in enabling GVCoT is the scarcity of image editing datasets with accurate edit region annotations. 
To overcome this, we develop a scalable multi-stage pipeline that automatically generates high-quality bounding boxes and segmentation masks for edited regions across diverse editing tasks.
We utilize this pipeline to construct \textbf{GVCoT-Edit-Instruct}, a large-scale dataset containing 1.8 million high-quality training samples. In particular, we adopt a progressive training recipe that combines supervised fine-tuning (SFT) and reinforcement learning (RL). The first phase focuses on equipping the model with foundational capabilities of drawing masks onto original images and producing structured visual reasoning chains before the image editing process. 
The second phase boosts both intermediate localization accuracy and final editing fidelity using Group Relative Policy Optimization (GRPO) \cite{liu2025flowgrpo}. 

While existing benchmarks such as ImgEdit \cite{ye2025imgedit} and GEdit-Bench \cite{liu2025step1x} primarily focus on object-salient scenes, they fall short in evaluating a model’s true spatial reasoning ability under complex editing scenarios. To address this, we introduce \textbf{SREdit-Bench}, a new benchmark comprising 590 carefully curated samples covering (1) non-object-salient and multiple entities scenes, and (2) fine-grained referring expressions in instructions. 
We evaluate 16 representative editing models and observe considerable performance gaps, highlighting the challenges of spatially grounded reasoning in image editing. We hope SpaEdit-Bench can serve as a new testbed for future research.

Our main contributions are summarized as follows:
\begin{itemize}
    \item We introduce GVCoT, a new image editing paradigm that integrates reasoning via visual intermediates, outperforming state-of-the-art approaches. 
    \item We develop a scalable curation pipeline and construct GVCoT-Edit-Instruct, a large-scale dataset comprising 1.8M high-quality pairs with region annotations.
    \item We propose a unified end-to-end training recipe that leverages progressive supervised fine-tuning and reinforcement learning with multi-dimensional rewards.
    \item We introduce SREdit-Bench, a new benchmark that assesses models' visual reasoning ability in image editing. Experiments demonstrate the superiority of our method.
\end{itemize}

\begin{figure*}[t]
    \centering
    \includegraphics[width=\linewidth]{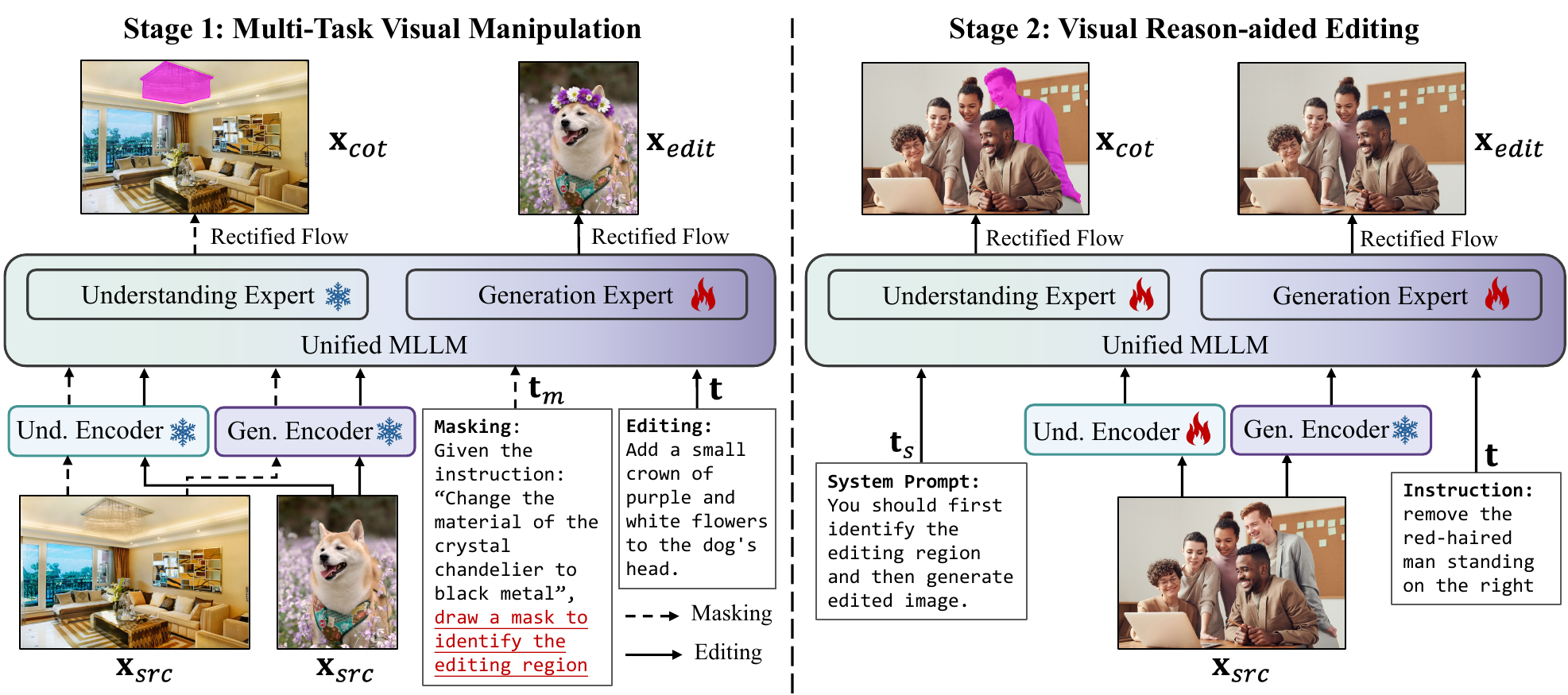}
    \vspace{-1.7em}
    \caption{\textbf{Supervised Fine-Tuning of our GVCoT training recipe.} Stage 1: Multi-Task Visual Manipulation, where the model's generation expert is trained in a multi-task setup to inject the newly masking skill. Stage 2: Visual Reason-aided Editing, where the entire model is trained to generate a faithful and interpretable visual reasoning image and then an edited image within a single sequence.}
    \vspace{-1.3em}
    \label{fig:method}
\end{figure*}

\section{Related Work}
\label{sec:related_work}

\noindent \textbf{Instruction-Guided Image Editing.} 
Diffusion models \cite{lipman2022flow,ho2020denoising,song2020score} have revolutionized visual content creation and manipulation. Early training-free works \cite{hertz2022p2p,meng2021sdedit,cao2023masactrl} modify content through latent inversion and attention-based controls. Training-based approaches \cite{wu2025qwenimagetechnicalreport,labs2025flux1kontextflowmatching,liu2025step1x,wang2025seededit,brooks2023instructpix2pix,geng2024instructdiffusion,zhang2025icedit} have shown strong capability by constructing high-quality training pairs. To handle more complex and compositional editing tasks, several approaches \cite{legoedit,camila,liu2025magicquill} employ an agentic scheme, where an MLLM first plans the instruction and then drives the diffusion process to execute sub-tasks. Additionally, several benchmarks \cite{jia2025compbench,yang2025complexedit,wang2025complexbench} evaluate model performance on complex tasks. CompBench \cite{jia2025compbench} features scenes that require sophisticated spatial and contextual reasoning, and Complex-Edit \cite{yang2025complexedit} progressively tests models by increasing instruction complexity.

\noindent \textbf{Multimodal Reasoning.}
The emergence of multimodal large language models \cite{bai2025qwen2,deng2025bagel,lin2025uniworld,ming_univision,song2025query,chen2025blip3o,xie2025show} has unlocked powerful multimodal reasoning capabilities. 
Prior works \cite{guo2025deepseek,huang2025vision} employ text CoT to enhance visual perception \cite{bai2025univg}, mathematical reasoning \cite{huang2025vision}, and visual generation \cite{duan2025got,fang2025got}. 
Unlikely, visual CoT integrates visual aids directly into the reasoning process. One approach uses external tools, \textit{e.g.} drawing auxiliary lines \cite{hu2024visualsketchpad}, zooming in \cite{su2025pixel,zheng2025deepeyes}, style transfer \cite{liu2025visual}, and sub-region highlighting \cite{fu2025refocus}. 
Another approach explores intrinsic visual CoT, where models generate visual thoughts natively \cite{cheng2025visual,shi2025mathcanvas,li2025imagine,chern2025thinking,li2025zebra}. Despite the promise, this approach is largely unexplored in image editing. Concurrently, MURE \cite{zou2025beyond} employs native interleaved CoT for image editing. However, it does not evaluate its spatial reasoning ability under complex tasks. 
\section{Method}

\begin{figure*}[ht]
    \centering
    \includegraphics[width=\linewidth]{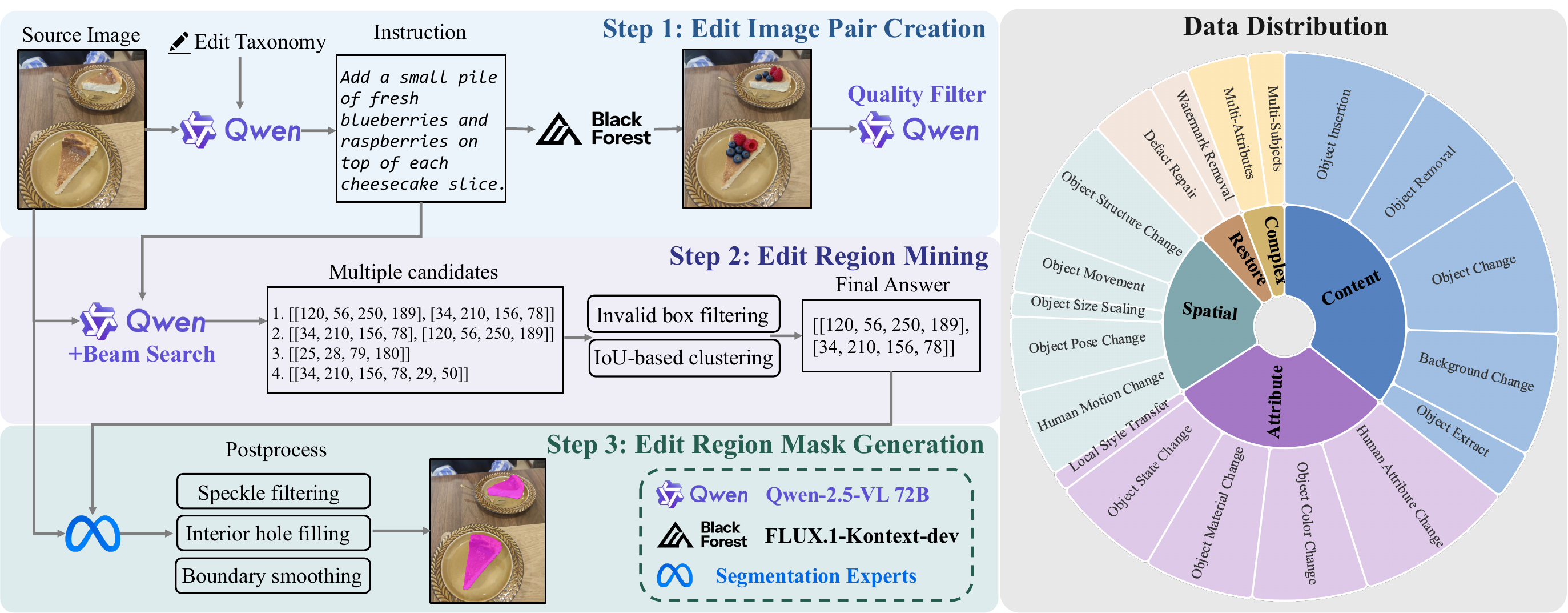}
    \vspace{-1.5em}
    \caption{\textbf{GVCoT-Edit-Instruct Data Pipeline.} Left: We design a scalable multi-stage data pipeline to curate high-quality samples with faithful editing region annotations, \textit{i.e.}, bounding boxes and masks. Right: The distribution of GVCoT-Edit-Instruct spanning 19 tasks.}
    \vspace{-1.3em}
    \label{fig:data}
\end{figure*}

\subsection{GVCoT Formulation}
Different from existing methods relying on textual intermediate reasoning results, our proposed GVCoT first infers an intermediate visual Chain-of-Thought (CoT) image and subsequently generates the final edited result.
Formally, given an input image $\mathbf{x}_{src} \in \mathbb{R}^{H \times W \times 3}$ and the editing instruction $\mathbf{t}$, the goal is to generate: (1) a visual thought map $\mathbf{x}_{cot} \in \mathbb{R}^{H \times W \times 3}$ that explicitly highlights editing regions, and (2) a final edited image $\mathbf{x}_{edit} \in \mathbb{R}^{H \times W \times 3}$. The overall process can be expressed as:
\vspace{-0.5em}
\begin{equation}
\mathbf{x}_{cot} = f_{\theta}(\mathbf{x}_{src}, \mathbf{t}), \quad 
\mathbf{x}_{edit} = f_{\theta}(\mathbf{x}_{src}, \mathbf{t}, \mathbf{x}_{cot}),
\end{equation}
where $f_{\theta}$ denotes the unified model.

\subsection{GVCoT Training Recipe}
We implement our GVCoT framework on Bagel \cite{deng2025bagel}, a unified model that has two distinct experts, an understanding expert and a generation expert. To stably internalize and improve the new visual reasoning skills without disrupting the model's original capability, we employ a two-phase training recipe: (1) Progressive Supervised Fine-tuning and (2) Reinforcement-based Refining.

\noindent \textbf{Progressive Supervised Fine Tuning.} 
The first phase aims to endow the model with the fundamental capability to generate an accurate and interpretable visual reasoning image $\mathbf{x}_{cot}$ before editing. We design a progressive strategy containing two steps, as shown in Fig.~\ref{fig:method}.

\textit{Step 1: Multi-Task Visual Manipulation.} This stage injects explicit spatial localization capability into the generation expert. To prevent catastrophic forgetting of prior editing skills, we adopt a multi-task objective: (1) masking: generating an image $\mathbf{x}_{cot}$ (which draws masks on the $\mathbf{x}_{src}$) based on source image $\mathbf{x}_{src}$ and masking instruction $\mathbf{t}_{m}$; (2) editing: predicting an edited image $\mathbf{x}_{edit}$ conditioned on $\mathbf{x}_{src}$ and edit instruction $T$. 
All images provided in the question are encoded into clean VAE and ViT tokens, serving as visual context. To preserve the model’s inherent reasoning abilities, we freeze the entire understanding expert and only train the generation expert as follows:
\begin{equation}
\lambda_{m}\mathcal{L}(\mathbf{x}_{cot}^{*}, f_{\theta}(\mathbf{x}_{src}, \mathbf{t}_{m})) + \lambda_{e}\mathcal{L}(\mathbf{x}_{edit}^{*}, f_{\theta}(\mathbf{x}_{src}, \mathbf{t}))
\end{equation}
where $\mathbf{x}_{cot}^{*}$ and $\mathbf{x}_{edit}^{*}$ indicates the ground-truth of visual reasoning and edit image, $\mathcal{L}$ is the flow matching loss \cite{liu2022flow,lipman2022flow}, and $\lambda_{m}$ and $\lambda_{e}$ are weights of two tasks to balance training dynamics.

\textit{Step 2: Visual Reason-aided Editing.} 
This stage aims to endow the model with reasoning-aware editing competence. The model is required to generate an intermediate visual reasoning image, and then the final edit step-by-step within a single sequence. Thus, the loss is:
\begin{equation}
    \mathcal{L}(\mathbf{x}_{cot}^{*}, f_{\theta}(\mathbf{t}_{s}, \mathbf{x}_{src}, \mathbf{t})) + \mathcal{L}(\mathbf{x}_{edit}^{*}, f_{\theta}(\mathbf{t}_{s}, \mathbf{x}_{src}, \mathbf{t}, \mathbf{x}_{cot}^{*}))
\end{equation}
where $\mathbf{t}_{s}$ is a predefined system text prompt. Unlike the first stage, all model components except the VAE encoder are unfrozen and trained jointly. Please refer to our Supplementary Material for more implementation details.

\noindent \textbf{Reinforcement-based Refining.} 
Then we aim to further refine the model's grounding accuracy and overall instruction following through reinforcement learning, \textit{i.e.}, Flow-GRPO \cite{liu2025flowgrpo}. 
However, jointly optimizing visual reasoning and final editing quality in a unified multi-task framework makes optimization unstable goal confusion. Thus, we adopt a progressive strategy, optimizing each generation step separately with tailored rewards.

\textit{Step 1: Visual Reasoning with Verified Rewards.} 
Low-quality visual reasoning may deteriorate the final result.
We enhance the localization accuracy of the model's visual thoughts using two verified rewards.
(1) Format Reward, which ensures the model follows a consistent reasoning–editing sequence rather than skipping or merging them. We train a binary classifier to distinguish whether an output image belongs to the visual thought stage or the editing stage. 
(2) IoU Reward, which measures the IoU between the ground-truth edit region mask and the predicted one. We extract the predicted mask by computing the pixel-wise difference between $\mathbf{x}_{src}$ and $\mathbf{x}_{cot}$.

\textit{Step 2: Editing with MLLM-as-a-Judge.} 
Even when using teacher-forcing visual thought to guide edits, the final results can still be inaccurate. To address this, we employ two rewards: (1) CoT-Edit Consistency Reward, which encourages the model to faithfully translate the teacher-forcing visual thought into accurate edits. (2) Image Quality Reward, which improves visual realism. Both rewards are quantified by MLLM-as-a-judge, leveraging the Qwen2.5-VL-72B \cite{bai2025qwen2} to generate a score. \textbf{More details on reward designs are provided in the supplementary}.

\begin{figure*}[t]
    \centering
    \includegraphics[width=\linewidth]{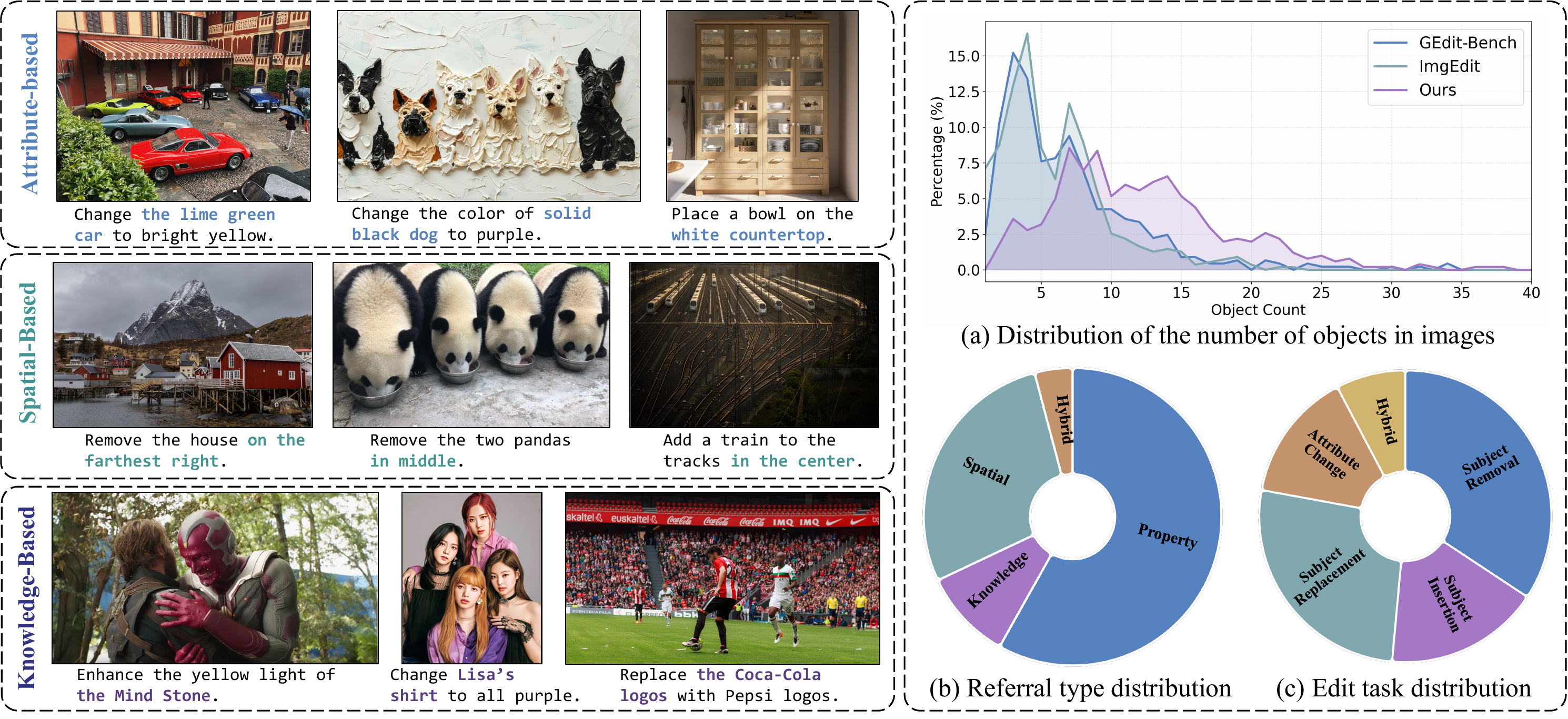}
    \vspace{-1.5em}
    \caption{\textbf{Illustration of the SREdit-Bench}. Left: We provide challenging scenarios featuring complex scenes and fine-grained referring expressions. Right: (a) We quantify scene complexity by counting editable objects and regions. Results show that SpaEdit-Bench concentrates on more sophisticated scenes than ImgEdit \cite{ye2025imgedit} and GEdit-Bench \cite{liu2025step1x}. (b) Referral type distribution. (c) Edit tasks distribution.}
    \vspace{-1.5em}
    \label{fig:benchmark}
\end{figure*}

\subsection{ GVCoT-Edit-Instruct Data Pipeline}
The major challenge is the lack of large-scale image editing training data with corresponding editing region annotations.
Thus, we design a scalable data construction pipeline and use it to create GVCoT-Edit-Instruct, comprising 1.8 million high-quality samples (see Fig. \ref{fig:data}). Each sample consists of a quadruple: a source image, an edit instruction, edit region annotations, and the target image. The pipeline consists of three main steps,  described below.

\noindent \textbf{Edit Image Pair Creation.}
We begin by constructing the source images, instructions, and edited images.
We collect 5.6 million images with at least $1K$ resolution from public datasets and websites, ensuring broad coverage of humans, objects, and scenes. 
We define a comprehensive edit taxonomy that spans diverse, real-world editing intents. Since our focus is on localized reasoning and editing, we exclude global edits such as style transfer and viewpoint change.
Guided by this taxonomy, Qwen2.5-VL \cite{bai2025qwen2} produces concise, natural user-style instructions, and FLUX.1 Kontext [Dev] \cite{labs2025flux1kontextflowmatching} synthesizes edited images. At last, an MLLM-based verifier filters out low-quality samples by measuring image naturalness and edit faithfulness. 

\noindent \textbf{Edit Region Mining.}
Then we mine the editing regions' annotations. Previous attempts \cite{ming_univision,lin2025uniworld} compute pixel differences between source and target images to acquire a regional mask. While this method is effective for rigid edits (\textit{e.g.}, color shift), it fails for flexible edits such as object motion or structural changes. 
We instead propose a more robust localization strategy. Qwen2.5-VL \cite{bai2025qwen2} predicts bounding box coordinates for the intended edit regions; multiple candidates are generated via beam search to minimize hallucinations. We filter out invalid boxes (\textit{e.g}, out-of-bounds, zero area, extreme aspect ratios) and perform IoU-based clustering to remove outliers. The averaged coordinates of the valid boxes are taken as the final results.

\noindent \textbf{Edit Region Mask Generation.}
At last, we generate a precise mask for each mined edit region. For object insertion, we directly use a box mask since the object boundary is unknown before editing. For modification and removal tasks, we leverage segmentation experts, \textit{i.e.}, SAM2 \cite{ravi2024sam2} and BiRefNet \cite{zheng2024birefnet}, to produce instance masks. Finally, a post-process is applied to fill the interior hole, remove exterior speckle, and smooth the boundaries.

\subsection{SREdit-Bench}
Existing benchmarks such as ImgEdit \cite{ye2025imgedit} and GEdit-Bench \cite{liu2025step1x} under-represent spatially complex editing scenarios, \textit{e.g.}, multiple similar editable entities, non-object-salient scenes, and tasks that demand fine-grained object referral. Some prior works \cite{jia2025compbench,yang2025complexedit,wang2025complexbench} consider multi-region editing and complex scenes; however, they do not target evaluating spatial reasoning ability and often rely on low-resolution images ($<$1024$\times$1024). To fill this gap, we introduce \textbf{SREdit-Bench}, a new benchmark focused on editing scenarios that require spatial reasoning.

\noindent \textbf{Benchmark Construction.} 
We curate a diverse set of high-quality source images ($>$1024$\times$1024) from the Internet ( e.g., Unsplash), and remove similar scenes to maximize diversity. To comprehensively evaluate models’ spatial reasoning in editing, we have two critical designs as shown in Fig. \ref{fig:benchmark}. The first is sophisticated scenes, including multiple entities and non-object-centric images. The second is fine-grained target referral in instruction, including three modes: (1) \emph{spatial}: explicit location or relational cues; (2) \emph{property}: appearance or attribute-based descriptions; and (3) \emph{knowledge}: implicit, context-dependent cues that require background or commonsense knowledge.

\noindent \textbf{Evaluation Protocol.} We use GPT4.1 \cite{gpt4} as an automated judge for consistent, scalable evaluation. Following VIEScore \cite{ku2023viescore}, we report three metrics: (1) SC (Semantic Consistency) — how well the edited result follows the instruction; (2) PQ (Perceptual Quality) — image naturalness and artifact presence; (3) O (Overall) — the geometric mean of SC and PQ, averaged across all samples.
\begin{table}[t]
\vspace{-0.9em}
\renewcommand\arraystretch{0.92}
\centering
\caption{\textbf{Quantitative results on SREdit-Bench.} $\text{SC}_g$, $\text{PQ}_g$, and $\text{O}_g$ indicate scores on Semantic Consistency, Perceptual Quality, and Overall. The best and second-best results are highlighted in bold and underlined, respectively.}
\resizebox{0.9\linewidth}{!}{
\begin{tabular}{lccc}
    \toprule
    \multirow{2}{*}{\textbf{Model}} & \multicolumn{3}{c}{\textbf{SREdit-Bench} $\uparrow$} \\
     \cmidrule(lr){2-4}
     & $\text{SC}_g$ & $\text{PQ}_g$ & $\text{O}_g$ \\
    \midrule
    \multicolumn{4}{c}{\textit{Product-level models}} \\
    \midrule
    GPT Image 1 [High] \cite{gptimage} & 9.02 & 8.42 & 8.56 \\
    FLUX.1 Kontext [Pro] & 8.69 & 8.40 & 8.41 \\
    Qwen-Image \cite{wu2025qwenimagetechnicalreport} & 8.57 & 8.43 & 8.32 \\
    \midrule
    \multicolumn{4}{c}{\textit{Generation Only}} \\
    \midrule
    Instruct-Pix2Pix \cite{brooks2023instructpix2pix} & 2.10 & 6.40 & 2.58 \\
    MagicBrush \cite{zhang2023magicbrush} & 2.99 & 5.91 & 3.42 \\
    ICEdit \cite{zhang2025icedit} & 4.41 & 7.71 & 4.81 \\
    OmniGen \cite{xiao2025omnigen} & 7.05 & 7.38 & 6.68 \\
    FLUX.1 Kontext [Dev] \cite{labs2025flux1kontextflowmatching} & 7.52 & 8.17 & 7.27 \\
    Step1X-Edit \cite{liu2025step1x} & 6.96 & 7.87 & 6.98 \\
    \midrule
    \multicolumn{4}{c}{\textit{Unified Understanding and Generation}} \\
    \midrule
    GoT \cite{fang2025got} & 5.78 & 7.59 & 5.83 \\
    Bagel \cite{deng2025bagel} & 8.02 & 7.90 & 7.75 \\
    Bagel-think \cite{deng2025bagel} & \underline{8.13} & 8.01 & \underline{7.82} \\
    UniWorld-v1 \cite{lin2025uniworld} & 5.92 & \underline{8.47} & 6.02 \\
    OmniGen2 \cite{wu2025omnigen2} & 6.52 & 7.54 & 6.44 \\
    Ming-UniVision \cite{ming_univision} & 6.05 & 6.85 & 5.96 \\
    \midrule
    \rowcolor{lightblue}\textbf{Bagel-GVCoT (Ours)} & \textbf{8.87} & \textbf{8.76} & \textbf{8.53} \\
    \rowcolor{lightblue}$\Delta$ Over Base Model & \textcolor{plusred}{+0.85} & \textcolor{plusred}{+0.86} & \textcolor{plusred}{+0.78} \\
    \bottomrule
\end{tabular}
}
\label{tab:sredit}
\end{table}

\begin{table}[t]
\centering
\caption{\textbf{Quantitative results in HumanEdit.} Our Bagel-GVCoT allowing native visual reasoning  outperforms previous approaches that rely on additional mask guidance.}
\vspace{-0.8em}
\resizebox{\linewidth}{!}{
\begin{tabular}{lcccc}
    \toprule
    \textbf{Method} & \textbf{CLIP-I}$\uparrow$ & \textbf{LPIPS}$\downarrow$ & \textbf{PSNR}$\uparrow$ & \textbf{SSIM}$\uparrow$ \\
    \midrule
    SmartEdit \cite{huang2024smartedit}  & 0.8841 & 0.2915 & 17.1728 & 0.6828 \\
    BrushNet \cite{ju2024brushnet} & 0.8986 & 0.1830 & 19.2172 & 0.7877 \\
    MagicQuill \cite{liu2025magicquill} & \underline{0.9381} & \underline{0.1162} & 22.2380 & \textbf{0.8981} \\
    MIND-Edit \cite{wang2025mind} & 0.9310 & 0.1245 & \underline{22.2714} & 0.8517 \\
    \midrule
    \textcolor{gray}{Bagel \cite{deng2025bagel}} & \textcolor{gray}{0.9124} & \textcolor{gray}{0.1721} & \textcolor{gray}{22.1843} & \textcolor{gray}{0.8640} \\ 
    \rowcolor{lightblue}\textbf{Bagel-GVCoT (Ours)} & \textbf{0.9451} & \textbf{0.1066} & \textbf{23.0161} & \underline{0.8943} \\
    \rowcolor{lightblue}$\Delta$ Over Base Model & \textcolor{plusred}{+0.0327} & \textcolor{plusred}{-0.0655} & \textcolor{plusred}{+0.8341} & \textcolor{plusred}{+0.0303} \\
    \bottomrule
\end{tabular}}
\label{tab:humanedit}
\end{table}

\begin{table*}[t]
    \renewcommand\arraystretch{0.95}
    \centering
    \caption{\textbf{Quantitative results on ImgEdit.} We use GPT-4.1 to evaluate all metrics. ``Overall'' is calculated by averaging all scores across tasks. The best and second-best results in open-sourced models are highlighted in \textbf{bold} and \underline{underlined}, respectively.}
    \vspace{-0.8em}
    \resizebox{\linewidth}{!}{
    \begin{tabular}{l|ccccccccc|c}
        \toprule
        \textbf{Model} & \bf Add & \bf Adjust & \bf Extract & \bf Replace & \bf Remove & \bf Background & \bf Style & \bf Hybrid & \bf Action & \bf Overall $\uparrow$ \\
        \midrule
        \multicolumn{11}{c}{\textit{Product-level models}} \\
        \midrule
        FLUX.1 Kontext [Pro] \cite{labs2025flux1kontextflowmatching} & 4.25 & 4.15 & 2.35 & 4.56 & 3.57 & 4.26 & 4.57 & 3.68 & 4.63 & 4.00 \\
        GPT Image 1 [High] \cite{gptimage} & 4.61 & 4.33 & 2.90 & 4.35 & 3.66 & 4.57 & 4.93 & 3.96 & 4.89 & 4.20 \\
        Qwen-Image \cite{wu2025qwenimagetechnicalreport} & 4.38 & 4.16 & 3.43 & 4.66 & 4.14 & 4.38 & 4.81 & 3.82 & 4.69 & 4.27 \\
        \midrule
        \multicolumn{11}{c}{\textit{Generation Only}} \\
        \midrule
        MagicBrush \cite{zhang2023magicbrush} & 2.84 & 1.58 & 1.51 & 1.97 & 1.58 & 1.75 & 2.38 & 1.62 & 1.22 & 1.90 \\
        Instruct-Pix2Pix \cite{brooks2023instructpix2pix} & 2.45 & 1.83 & 1.44 & 2.01 & 1.50 & 1.44 & 3.55 & 1.20 & 1.46 & 1.88 \\
        AnyEdit \cite{yu2025anyedit} & 3.18 & 2.95 & 1.88 & 2.47 & 2.23 & 2.24 & 2.85 & 1.56 & 2.65 & 2.45 \\
        UltraEdit \cite{zhao2024ultraedit} & 3.44 & 2.81 & 2.13 & 2.96 & 1.45 & 2.83 & 3.76 & 1.91 & 2.98 & 2.70 \\
        OmniGen \cite{xiao2025omnigen} & 3.47 & 3.04 & 1.71 & 2.94 & 2.43 & 3.21 & 4.19 & 2.24 & 3.38 & 2.96 \\
        ICEdit \cite{zhang2025icedit} & 3.58 & 3.39 & 1.73 & 3.15 & 2.93 & 3.08 & 3.84 & 2.04 & 3.68 & 3.05 \\
        Step1X-Edit \cite{liu2025step1x} & 3.88 & 3.14 & 1.76 & 3.40 & 2.41 & 3.16 & 4.63 & 2.64 & 2.52 & 3.06 \\
        FLUX.1 Kontext [Dev] \cite{labs2025flux1kontextflowmatching} & \textbf{4.12} & \underline{3.80} & 2.04 & \underline{4.22} & 3.09 & 3.97 & 4.51 & \textbf{3.35} & 4.25 & \underline{3.71} \\
        \midrule
        \multicolumn{11}{c}{\textit{Unified Understanding and Generation}} \\
        \midrule
        GoT \cite{fang2025got} & 3.74 & 3.06 & 1.33 & 2.72 & 2.46 & 2.33 & 3.45 & 1.77 & 2.50 & 2.65 \\
        Bagel \cite{deng2025bagel} & 3.56 & 3.31 & 1.70 & 3.3 & 2.62 & 3.24 & 4.49 & 2.38 & 4.17 & 3.20 \\
        Bagel-think \cite{deng2025bagel} & 3.65 & 3.53 & 2.03 & 3.60 & 3.03 & 3.45 & 4.43 & 2.59 & 4.22 & 3.39 \\
        UniWorld-V1 \cite{lin2025uniworld} & 3.82 & 3.64 & 2.27 & 3.47 & 3.24 & 2.99 & 4.21 & \underline{2.96} & 2.74 & 3.26 \\
        OmniGen2 \cite{wu2025omnigen2} & 3.57 & 3.06 & 1.77 & 3.74 & 3.20 & 3.57 & \textbf{4.81} & 2.52 & \textbf{4.68} & 3.44 \\
        Ming-UniVision \cite{ming_univision} & 3.55 & 3.14 & 1.52 & 3.25 & \underline{3.29} & 2.77 & 3.99 & 2.74 & 3.91 & 3.06 \\
        BLIP3o-NEXT \cite{chen2025blip3o} & 4.00 & 3.78 & \underline{2.39} & 4.05 & 2.61 & \textbf{4.30} & \underline{4.64} & 2.67 & 4.13 & 3.62 \\
        \midrule
        \rowcolor{lightblue}\textbf{Bagel-GVCoT (Ours)} & \underline{4.02} & \textbf{4.07} & \textbf{2.92} & \textbf{4.23} & \textbf{3.74} & \underline{4.16} & 3.83 & 2.82 & \underline{4.48} & \textbf{3.82}  \\
        \rowcolor{lightblue}$\Delta$ Over Base Model & \textcolor{plusred}{+0.46} & \textcolor{plusred}{+0.76} & \textcolor{plusred}{+1.22} & \textcolor{plusred}{+0.93} & \textcolor{plusred}{+1.12} & \textcolor{plusred}{+0.92} & \textcolor{plusred}{-0.66} & \textcolor{plusred}{+0.44} & \textcolor{plusred}{+0.31} & \textcolor{plusred}{+0.62} \\
        \bottomrule
    \end{tabular}
    }
    \label{tab:imgedit}
\end{table*}

\section{Experiments}

\subsection{Main Results}
\noindent \textbf{Comparison with general image editing methods.}
We first compare our Bagel-GVCoT against 17 prominent general image editing algorithms, including top-performing product-level models FLUX.1 Kontext Pro \cite{labs2025flux1kontextflowmatching}, Qwen-Image \cite{wu2025qwenimagetechnicalreport}, GPT Image 1 \cite{gptimage}, and powerful open-source methods, including the pure generation models \cite{zhang2023magicbrush,brooks2023instructpix2pix,yu2025anyedit,zhao2024ultraedit,xiao2025omnigen,zhang2025icedit,liu2025step1x,wu2025qwenimagetechnicalreport} and unified models \cite{lin2025uniworld,wu2025omnigen2,ming_univision,chen2025blip3o}. 

Tab.~\ref{tab:sredit} shows results on our SREdit-Bench. Bagel-GVCoT achieves the highest overall performance under challenging image-editing scenarios with multiple objects and complex spatial relations. It attains an overall score of 8.53, surpassing the diffusion specialists Qwen-Image \cite{wu2025qwenimagetechnicalreport}. Moreover, our Bagel-GVCoT outperforms text-CoT-based reasoning models, including GoT \cite{fang2025got} and Bagel-think \cite{deng2025bagel}, demonstrating the superiority of incorporating explicit visual reasoning into the editing process.

We also evaluate Bagel-GVCoT across various sub-tasks on ImgEdit \cite{ye2025imgedit}. As listed in Tab. \ref{tab:imgedit}, our method achieves the best overall performance among open-source models, second only to Qwen-Image \cite{wu2025qwenimagetechnicalreport}. Although the model attains a relatively lower score (3.83) on the style transfer task, this represents a reasonable trade-off—our framework is specifically optimized for precise spatial reasoning and localized editing rather than global stylistic manipulation. We provide more results in the supplementary.

\noindent \textbf{Comparison with mask-based image editing methods.} 
Since our method generates intermediate spatial cues to guide the subsequent editing process, we also compare it with mask-based editing models \cite{ju2024brushnet,huang2024smartedit,liu2025magicquill,wang2025mind,qu2025vincie} that rely on input masks as external spatial guidance. We follow the evaluation setup of HumanEdit \cite{bai2024humanedit} in \cite{wang2025mind} for a fair comparison. Tab. \ref{tab:humanedit} demonstrates that our Bagel-GVCoT outperforms all mask-based counterparts. As the base model Bagel \cite{deng2025bagel} fails to surpass these methods, it further validates the strength of our visual reasoning paradigm.

\begin{figure}
    \centering
    \includegraphics[width=\linewidth]{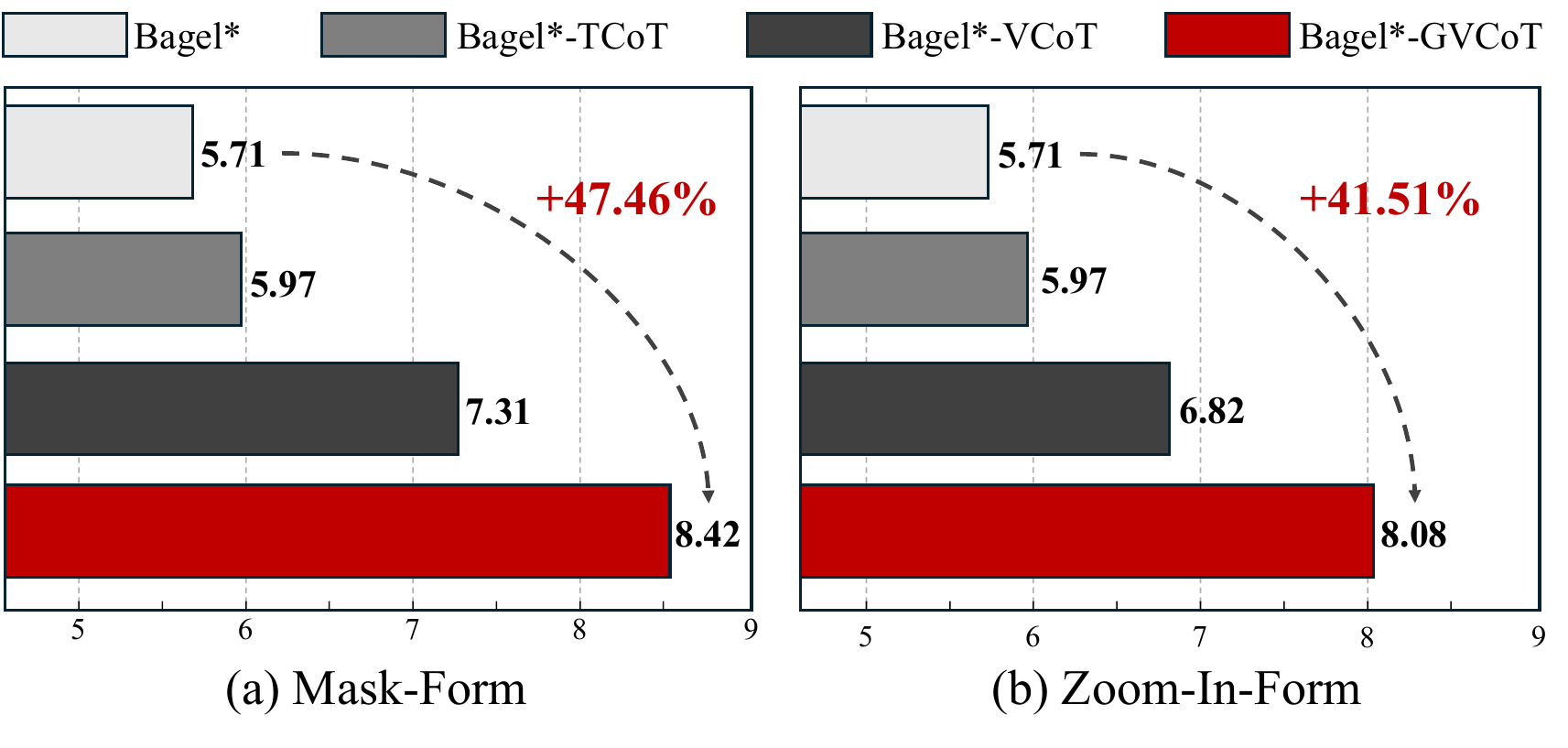}
    \vspace{-1.7em}
    \caption{\textbf{Quantitative comparison of different visual reasoning paradigms on SREdit-Bench.} The performance is measured by $\text{O}_g$. Under both visual cue forms, our method consistently surpasses the text CoT and Visual CoT with considerable margins. }
    \vspace{-1.3em}
    \label{fig:ablation_visual}
\end{figure}
\begin{figure*}[t]
    \centering
    \includegraphics[width=\linewidth]{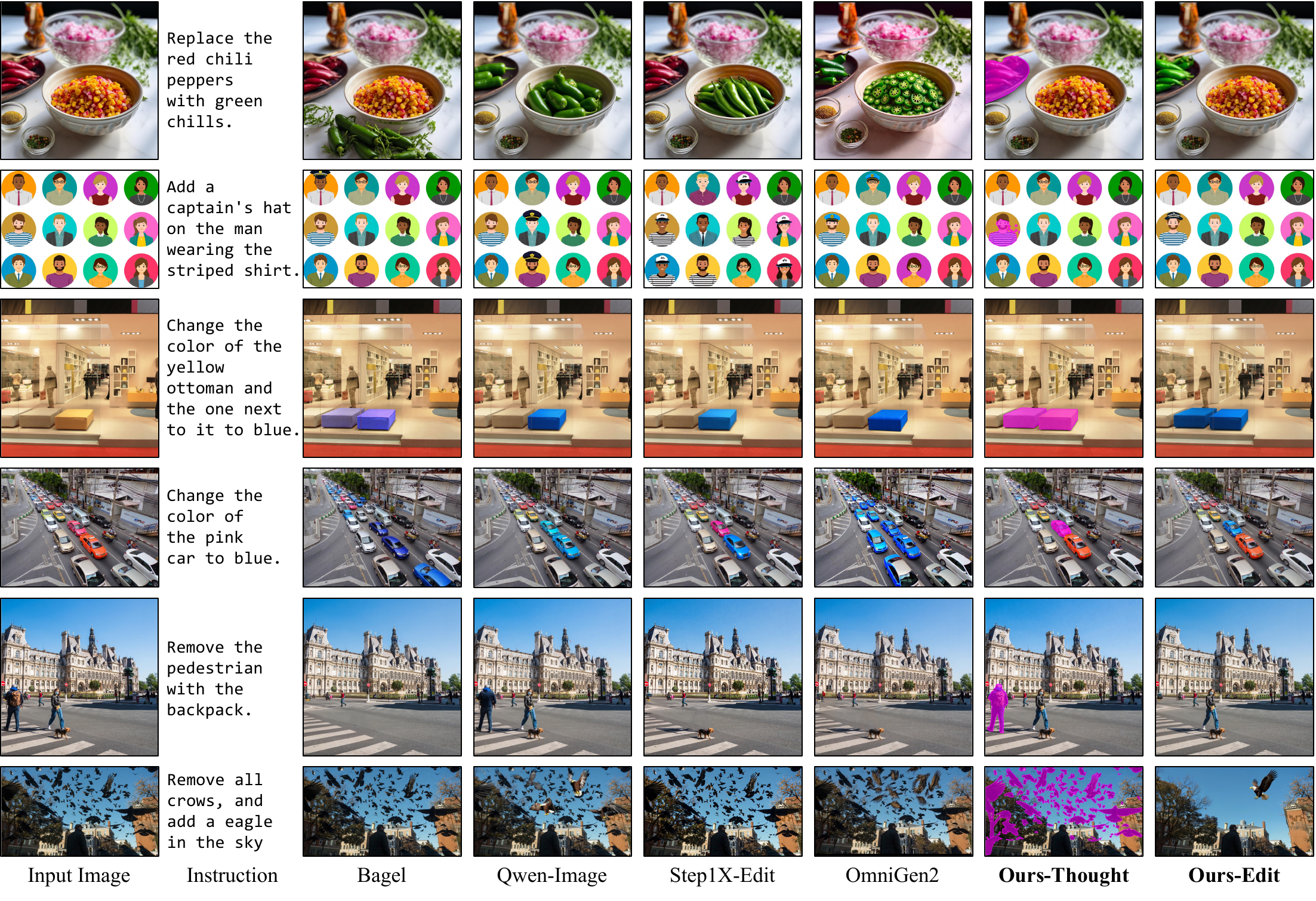}
    \vspace{-2.0em}
    \caption{\textbf{Qualitative comparison in SREdit-Bench.} Our method demonstrates superior spatial reasoning and instruction adherence compared to existing open-source models, especially when handling complex, multi-object editing tasks. }
    \vspace{-1em}
    \label{fig:results}
\end{figure*}

\subsection{Ablation Studies}

\noindent \textbf{Comparison of Visual Reasoning Paradigms.}
We compare two distinct visual reasoning paradigms: Visual CoT (VCoT), which relies on external tools, and our Generative Visual CoT (GVCoT), which produces reasoning cues in an end-to-end generative manner. To comprehensively evaluate, we design two forms of visual thought that reflect how spatial reasoning is represented and applied during editing:
\begin{itemize}
    \item \textbf{Mask-Form}: In VCoT, the model first predicts bounding box coordinates and then employs SAM2 \cite{ravi2024sam2} to generate a segmentation mask, which is fused with the input image. In contrast, GVCoT directly generates the mask onto the image through a generative process.
    \item \textbf{Zoom-In-Form}: VCoT predicts the bounding box and then crops the corresponding region from the input image, whereas GVCoT generates a zoomed-in sub-image.
\end{itemize}
We include control group: Bagel\textsuperscript{*}, (the base model directly fine-tuned on our dataset), and Bagel with text CoT (Bagel\textsuperscript{*}-TCoT), which only generates textual coordinates for fair comparison. The comparative results under two visual cue forms are illustrated in Fig.~\ref{fig:ablation_visual}, showing that Bagel\textsuperscript{*}-GVCoT consistently outperforms Bagel\textsuperscript{*}-VCoT and Bagel\textsuperscript{*}-TCoT with considerable margins.
Our GVCoT paradigm are more effective than one rely on external tools.

\begin{table}[t]
\renewcommand\arraystretch{0.98}
\centering
\caption{\textbf{Comparative results of GVCoT and Visual CoT using SREdit-Bench.} TF-thought denotes skipping the reasoning step by using teacher-forcing visual thought.}
\vspace{-0.7em}
\resizebox{0.97\linewidth}{!}{
\begin{tabular}{llcccc}
    \toprule
     & \textbf{Model} & \textbf{IoU}$\uparrow$ & \textbf{$\text{SC}_g$}$\uparrow$ & \textbf{$\text{PQ}_g$}$\uparrow$ & \textbf{$\text{O}_g$}$\uparrow$ \\
    \midrule
        & \textcolor{gray}{Bagel\textsuperscript{*}} & -- & \textcolor{gray}{6.21} & \textcolor{gray}{6.09} & \textcolor{gray}{5.71} \\
        & \textcolor{gray}{TCoT} & -- & \textcolor{gray}{6.54} & \textcolor{gray}{6.22} & \textcolor{gray}{5.97} \\
    \midrule
    \multirow{4}{*}{\rotatebox{90}{Mask}} & VCoT & 0.68 & 7.55 & 7.92 & 7.31 \\
    & \textbf{GVCoT} & 0.60 & 8.53 & 8.62 & 8.42 \\
    & \cellcolor{lightblue}{VCoT w/ TF-thought} & \cellcolor{lightblue}{--} & \cellcolor{lightblue}{8.03} & \cellcolor{lightblue}{7.96} & \cellcolor{lightblue}{7.75} \\
    & \cellcolor{lightblue}{\textbf{GVCoT w/ TF-thought}} & \cellcolor{lightblue}{--} & \cellcolor{lightblue}{8.79} & \cellcolor{lightblue}{8.95} & \cellcolor{lightblue}{8.72} \\
    \midrule
    \multirow{4}{*}{\rotatebox{90}{Zoom-In}} & VCoT & -- & 7.05 & 7.14 & 6.82 \\
    & \textbf{GVCoT} & -- & 8.12 & 8.30 & 8.08 \\
    \cdashline{2-6}
    & \cellcolor{lightblue}{VCoT w/ TF-thought} & \cellcolor{lightblue}{--} & \cellcolor{lightblue}{7.78} & \cellcolor{lightblue}{7.67} & \cellcolor{lightblue}{7.54} \\
    & \cellcolor{lightblue}{\textbf{GVCoT w/ TF-thought}} & \cellcolor{lightblue}{--} & \cellcolor{lightblue}{8.39} & \cellcolor{lightblue}{8.58} & \cellcolor{lightblue}{8.33} \\
    \bottomrule
\end{tabular}}
\vspace{-1em}
\label{tab:ablation_visalcot}
\end{table}

\noindent \textbf{In-depth Analysis of GVCoT and VCoT.} 
To understand why our GVCoT outperforms VCoT, we conduct an in-depth analysis, as shown in Tab.~\ref{tab:ablation_visalcot}. First, we measure the localization accuracy of visual thoughts by computing IoU with ground-truth masks. The two paradigms exhibit similar localization ability (0.68 for VCoT vs. 0.66 for GVCoT). 
Second, to isolate the impact of visual thought accuracy to edit results, we utilize teacher-forcing visual thoughts (denoted as TF-thought) for two paradigms. Results show GVCoT achieves a significantly higher $\text{O}_g$ score (8.72) compared to VCoT (7.75). GVCoT demonstrates stronger thought–edit consistency and the trend holds across both Mask and Zoom-In cue forms. 
Given the similar IoU but large performance gap in editing quality, the advantage of GVCoT clearly lies not in localization precision, but in how effectively it leverages spatial information. 

\noindent \textbf{Effectiveness of supervised fine-tuning designs.} 
Tab.~\ref{tab:ablation_sft} shows the ablation results of our two-stage SFT. Step~1 primarily enhances localization capability, Step~2 improves thought-editing consistency and editing quality. Combining both stages leads to the best overall performance.

\begin{table}[t]
\renewcommand\arraystretch{0.95}
\centering
\caption{\textbf{Ablation results on progressive SFT.}}
\vspace{-0.5em}
\resizebox{0.87\linewidth}{!}{
\begin{tabular}{ll|cccc}
    \toprule
    \textbf{Step 1} & \textbf{Step 2} & \textbf{IoU}$\uparrow$ & \textbf{$\text{SC}_g$}$\uparrow$ & \textbf{$\text{PQ}_g$}$\uparrow$ & \textbf{$\text{O}_g$}$\uparrow$ \\
    \midrule
    $\checkmark$ &  & 0.64 & 7.92 & 8.01 & 7.75 \\
     & $\checkmark$ & 0.61 & 8.21 & 8.33 & 8.10 \\
    $\checkmark$ & $\checkmark$ & \textbf{0.66} & \textbf{8.53} & \textbf{8.62} & \textbf{8.42} \\
    \bottomrule
\end{tabular}}
\vspace{-2em}
\label{tab:ablation_sft}
\end{table}

\noindent \textbf{Effectiveness of reinforcement learning designs.}
In Tab. \ref{tab:ablation_rl}, we perform ablation studies on our progressive reinforcement learning with multi-reward designs. Firstly, the incorporation of $\text{RL}$ boosts performance ($\text{O}_g$ from 8.42 to 8.53) and enhances spatial accuracy (IoU from 0.60 to 0.67). 
Secondly, the multi-stage setup is effective, as removing Stage 2 substantially degrades results. 
Thirdly, two rewards for improving the localization accuracy of visual thought prove vital, with their removal drops the IoU score, e.g., from 0.67 to 0.50 and 0.62, respectively.
Finally, both the CoT-Edit consistency reward and the Image quality reward contribute positively to ensuring both fidelity and the faithful translation of the visual thought into the final edit.

\begin{table}[t]
\renewcommand\arraystretch{0.95}
\centering
\caption{\textbf{Ablation studies on Reinforcement Learning (RL).} We analyze (1) multi-stage training strategy, (2) visual-thought reward design, and (3) editing reward design.}
\vspace{-0.5em}
\resizebox{\linewidth}{!}{
\begin{tabular}{lcccc}
    \toprule
    \textbf{Setting} & \textbf{IoU}$\uparrow$ & \textbf{$\text{SC}_g$}$\uparrow$ & \textbf{$\text{PQ}_g$}$\uparrow$ & \textbf{$\text{O}_g$}$\uparrow$ \\
    \midrule
    SFT Only  & 0.60 & 8.53 & 8.62 & 8.42 \\
    \textbf{SFT+RL} & \textbf{0.67} & \textbf{8.57} & \textbf{8.76} & \textbf{8.53} \\
    \midrule
    \multicolumn{5}{l}{\textcolor{gray}{\textbf{(1) Multi-stage RL training}}} \\
    \quad w/o Stage~2 & 0.67 & 8.32 & 8.47 & 8.25 \\
    \midrule
    \multicolumn{5}{l}{\textcolor{gray}{\textbf{(2) Visual thought reward design}}} \\
    Full reward set & \textbf{0.67} & \textbf{8.32} & \textbf{8.47} & \textbf{8.25} \\
    \quad w/o IoU reward & 0.50 & 8.32 & 8.39 & 8.13 \\
    \quad w/o Format reward & 0.62 & 8.13 & 8.24 & 8.19 \\
    \midrule
    \multicolumn{5}{l}{\textcolor{gray}{\textbf{(3) Editing reward design}}} \\
    Full reward set & \textbf{0.67} & \textbf{8.57} & \textbf{8.76} & \textbf{8.53} \\
    \quad w/o CoT-Edit consistency & 0.67 & 8.49 & 8.63 & 8.45 \\
    \quad w/o Image quality reward & 0.67 & 8.45 & 8.75 & 8.49 \\
    \bottomrule
\end{tabular}}
\vspace{-1em}
\label{tab:ablation_rl}
\end{table}

\section{Conclusion}
We introduce the Generative Visual Chain-of-Thought (GVCoT) framework, designed to endow unified models with intrinsic spatial reasoning capabilities for image editing. Leveraging our curated large-scale dataset, GVCoT-Edit-Instruct, which contains 1.8 million high-quality editing images with detailed region annotations, we adopt a two-phase training recipe to develop Bagel-GVCoT. 
This enables the model to accurately ground instructions and effectively handle complex image editing scenarios, including sophisticated scenes, intricate spatial relationships, and fine-grained object referring. 
Results on our SREdit-Bench benchmark demonstrate that Bagel-GVCoT achieves a 47.46\% relative improvement over the baseline. 
Crucially, we find that the generative visual reasoning way can more effectively exploit the spatial signals than the agentic one, which needs external tools or models to produce them. We hope this work will serve as a robust baseline for tackling complex and challenging image editing tasks.

\noindent \textbf{Acknowledgment.} This work was supported by the National Nature Science Foundation of China (Grant 62476029, 62225601, U23B2052), funded by the Fundamental Research Funds for the Beijing University of Posts and Telecommunications under Grant 2025TSQY08, the Beijing Natural Science Foundation Project No. L242025, the BUPT Excellent Ph.D. Students Foundation No. CX20242081, and sponsored by Beijing Nova Program and the Beijing Key Laboratory of Multimodal Data Intelligent Perception and Governance.

{
    \small
    \bibliographystyle{ieeenat_fullname}
    \bibliography{main}
}


\end{document}


\title{Generative Visual Chain-of-Thought for Image Editing}
\maketitlesupplementary

\begin{table}[t]
    \centering
    \caption{\textbf{Quantitative results on GEdit-Bench} We use Semantic Consistency (G\_SC), Perceptual Quality (G\_PQ), and Overall Score (G\_O) metrics measured by GPT 4.1.}
    \resizebox{\linewidth}{!}{
    \begin{tabular}{lccc}
    \toprule
    \multirow{2}{*}{\bf Model} & \multicolumn{3}{c}{\bf GEdit-Bench-EN (Full set)}
    \\
    \cmidrule{2-4}
    & G\_SC$\uparrow$ & G\_PQ$\uparrow$ & G\_O$\uparrow$ \\
    \midrule
    \multicolumn{4}{c}{\textit{Product-level models}} \\
    \midrule
    FLUX.1 Kontext [Pro]~\citep{labs2025flux1kontextflowmatching} & 7.02 & 7.60 & 6.56 \\
    Gemini 2.0~\citep{googleGemini2} & 6.73 & 6.61 & 6.32 \\
    Gemini 2.5 Flash Image Preview~\citep{googlegemini2_5} & 7.28 & 7.83 & 6.93 \\
    Gemini 2.5 Flash Image~\citep{googlegemini2_5} & 7.41 & 7.96 & 7.10 \\
    GPT-Image-1 [High]~\citep{gptimage} & 7.85 & 7.62 & 7.53 \\
    Qwen-Image-Edit-2509~\citep{wu2025qwenimagetechnicalreport} & 8.15 & 7.86 &  7.54 \\
    Qwen-Image-Edit~\citep{wu2025qwenimagetechnicalreport} & 8.00 & 7.86 & 7.56 \\
    \midrule
    \multicolumn{4}{c}{\textit{Generation Only}} \\
    \midrule
    AnyEdit~\citep{yu2025anyedit} & 3.18 & 5.82 & 3.21 \\
    Instruct-Pix2Pix \citep{brooks2023instructpix2pix} & 3.58 & 5.49 & 3.68 \\
    MagicBrush~\citep{zhang2023magicbrush} & 4.68 & 5.66 & 4.52 \\
    OmniGen~\citep{xiao2025omnigen} & 5.96 & 5.89 & 5.06 \\
    FLUX.1 Kontext [Dev]~\citep{labs2025flux1kontextflowmatching} & 6.52 & 7.38 & 6.00 \\
    Lego-Edit~\citep{legoedit} & 5.99 & 7.45 & 6.64 \\
    Step1X-Edit~\citep{liu2025step1x} & 7.66 & 7.35 & 6.97 \\
    \midrule
    \multicolumn{4}{c}{\textit{Unified Understanding and Generation}} \\
    \midrule
    UniWorld-v1~\citep{lin2025uniworld} & 4.93 & 7.43 & 4.85 \\
    OmniGen2~\citep{wu2025omnigen2} & 7.16 & 6.77 & 6.41 \\
    Bagel~\citep{deng2025bagel} & 7.36 & 6.83 & 6.52 \\
    Ming-Univision~\citep{ming_univision} & 6.04 & 6.86 & 5.54 \\
    \midrule
    \rowcolor{lightblue}\textbf{Bagel-NVCoT (Ours)} & 7.53 & 7.01 & 6.58 \\
    \rowcolor{lightblue}$\Delta$ Over Base Model & \textcolor{plusred}{+0.37} & \textcolor{plusred}{+0.024} & \textcolor{plusred}{+0.17} \\
    \bottomrule
    \end{tabular}}
\label{tab:gedit}
\end{table}

\section{Method Details}
\subsection{Reward Models}
In our reinforcement learning phase, we use MLLM-as-a-Judge to further improve the semantic consistency and image quality of the final edited results. We have two rewards based on Qwen2.5-VL-72B \cite{bai2025qwen2}: 1) CoT-Edit Consistency Reward, which encourages the model to translate the teacher-forcing visual thought into accurate edits faithfully. The detailed prompt template is displayed in Tab. \ref{tab:eval_gvcot_prompt}. 2) Image Quality Reward, which improves visual realism. The detailed prompt template is shown in Tab. \ref{tab:eval_pq_prompt}.

\subsection{Implementation Details}
We implement our framework on top of the publicly available BAGEL-7B-MoT \cite{deng2025bagel} model. All training experiments were conducted on a cluster of 64 NVIDIA H20 GPUs. We use the AdamW \cite{loshchilov2017decoupled} optimizer for both training stages. The detailed hyperparameters for our two-stage training in our Progressive Supervised Fine-Tuning, corresponding to Stage 1 (\textit{Multi-Task Visual Manipulation}) and Stage II (\textit{Visual Reason-aided Editing}), are provided in Tab. \ref{tab:hyperparams_sft}. The detailed hyperparameters for two stages in our Reinforcement-based Refining, corresponding to Stage 1 (\textit{Visual Reasoning with Verified Rewards}) and Stage 2 (\textit{Editing with MLLM-as-a-Judge}), are provided in Tab. \ref{tab:hyperparams_rl}.

\section{Limitations}
\begin{figure}
    \centering
    \includegraphics[width=\linewidth]{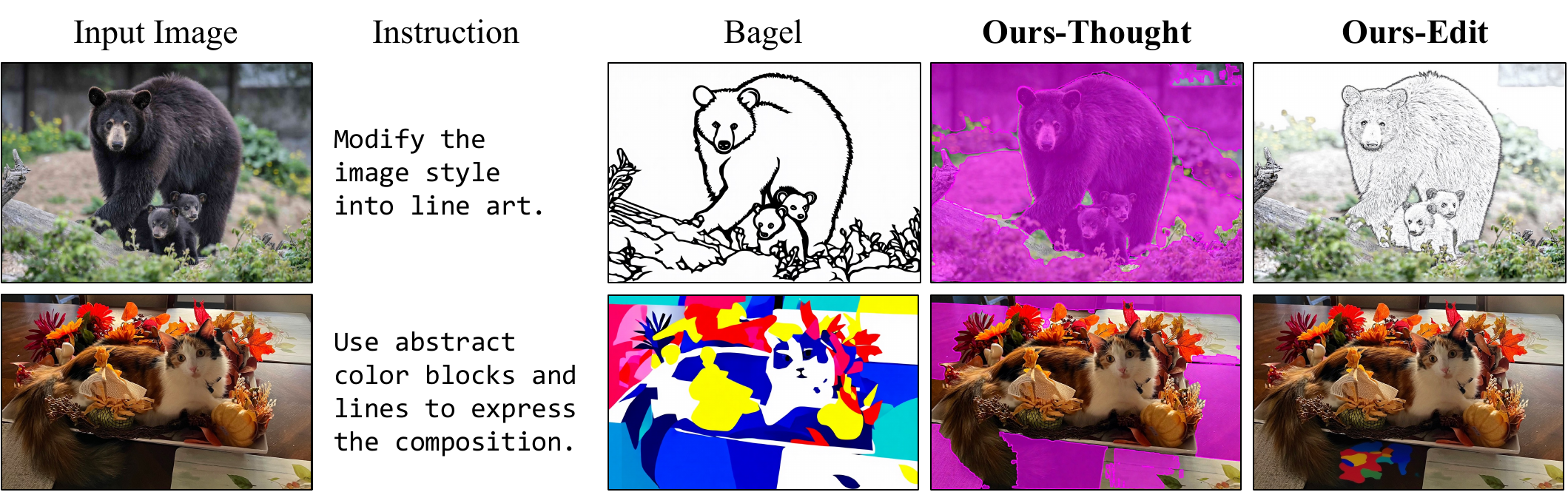}
    \vspace{-1.3em}
    \caption{\textbf{Limitation of our Bagel-GVCoT.} Our approach constrains under style transfer tasks compared with the baseline.}
    \vspace{-1.5em}
    \label{fig:result_limitation}
\end{figure}

Our GVCoT is primarily designed for spatially grounded, localized image editing, and thus our training dataset does not involve global style transfer samples. Our method's behavior on global style transfer tasks can be less competitive; the qualitative results are shown in Fig. \ref{fig:result_limitation}. 
The model may not naturally identify the entire image as the intended edit region, and its stylistic transformation may be less pronounced compared to the base model Bagel \cite{deng2025bagel}.
We do not view this as a fundamental limitation, as our core objective is to strengthen spatial reasoning and precise local manipulation in complex scenes. Fully addressing global style editing would likely require mechanisms beyond spatial localization, which we leave for future work.

\begin{table*}[ht]
\centering
\caption{\textbf{Quantitative results on Emu Edit and MagicBrush test sets.} All results except Bagel \cite{deng2025bagel} and our Bagel-GVCoT are directly from Unireal \cite{chen2025unireal}. The best and second-best results are highlighted in \textbf{bold} and \underline{underlined}, respectively.}
\vspace{-0.5em}
\label{tab:magicbrush}
\resizebox{\linewidth}{!}{
\centering
\begin{tabular}{lccccc|ccccc}
    \toprule
     & \multicolumn{5}{c|}{\textbf{Emu Edit Test set}} & \multicolumn{5}{c}{\textbf{MagicBrush Test Set}}\\
    \midrule
    Method & $\text{CLIP}_{dir}\!\uparrow$ &  $\text{CLIP}_{im}\!\uparrow$ & $\text{CLIP}_{out}\!\uparrow$ &  $\text{L1}\!\downarrow$  & DINO$\uparrow$  & $\text{CLIP}_{dir}\!\uparrow$ &  $\text{CLIP}_{im}\!\uparrow$ & $\text{CLIP}_{out}\!\uparrow$ &  $\text{L1}\!\downarrow$ & DINO$\uparrow$  \\
    \midrule
    InstructPix2Pix~\cite{brooks2023instructpix2pix} & 0.078 & 0.834 & 0.219 & 0.121 & 0.762 
                    & 0.115 & 0.837 & 0.245 & 0.093 & 0.767 \\
    MagicBrush~\cite{zhang2023magicbrush}     & 0.090 & 0.838 & 0.222 & 0.100 & 0.776 
                    & 0.123 & 0.883 & 0.261 & 0.058 & 0.871 \\
    PnP~\cite{tumanyan2023pnp}            & 0.028 & 0.521 & 0.089 & 0.304 & 0.153 
                    & 0.025 & 0.568 & 0.101 & 0.289 & 0.220 \\
    Null-Text Inv.~\cite{mokady2023null} & 0.101 & 0.761 & 0.236 & 0.075 & 0.678 
                    & 0.121 & 0.752 & 0.263  & 0.077 & 0.664 \\
    
    UltraEdit~\cite{zhao2024ultraedit}      & 0.107 & 0.793 & 0.283 & 0.071 & \underline{0.844} 
                    & -  & 0.868  & -  & \textbf{0.088}  & 0.792 \\
                    
    Emu Edit~\cite{sheynin2024emu}       &0.109 & 0.859 & 0.231 & 0.094 & 0.819
                    &0.135 & 0.897 & 0.261 & 0.052 & \underline{0.879} \\
    ACE~\cite{han2024ace}      & 0.086 & \textbf{0.895}  &  0.274  &  0.076 &  \textbf{0.862}
                    & -  & -  & 0.284  & -  & - \\
    
    OmniGen~\cite{xiao2025omnigen}      & -  & 0.836 &  0.233 & -  & 0.804
                    & -  &- & -  & -  & - \\
    
    PixWizard~\cite{lin2024pixwizard}      & 0.104  & 0.845  & 0.248  & 0.069  & 0.798 
                    & 0.124  & 0.884  & 0.265  & 0.063  & 0.876 \\
    
    Unireal~\cite{chen2025unireal}  & 0.127 & 0.851 & 0.285 & \textbf{0.099} & 0.790 & 0.151 & 0.903 & 0.308 & \underline{0.081} & 0.837\\
    Bagel \cite{deng2025bagel} & \underline{0.132} & 0.858 & \underline{0.290} & 0.085 & 0.810 & \underline{0.155} & \underline{0.908} & \textbf{0.312} & 0.075 & 0.852 \\
    \midrule
    \rowcolor{lightblue}\textbf{Bagel-GVCoT (Ours)} & \textbf{0.141} & \underline{0.867} & \textbf{0.301} & \underline{0.092} & 0.835 & \textbf{0.165} & \textbf{0.915} & \underline{0.310} & 0.080 & \textbf{0.883} \\
    \rowcolor{lightblue}$\Delta$ Over Base Model & \textcolor{plusred}{+0.009} & \textcolor{plusred}{+0.009} & \textcolor{plusred}{+0.011} & \textcolor{plusred}{+0.007} & \textcolor{plusred}{+0.025} & \textcolor{plusred}{+0.010} & \textcolor{plusred}{+0.007} & \textcolor{plusred}{-0.002} & \textcolor{plusred}{+0.005} & \textcolor{plusred}{+0.031} \\
    \bottomrule
\end{tabular}
}
\end{table*}

\begin{figure*}
    \centering
    \includegraphics[width=\linewidth]{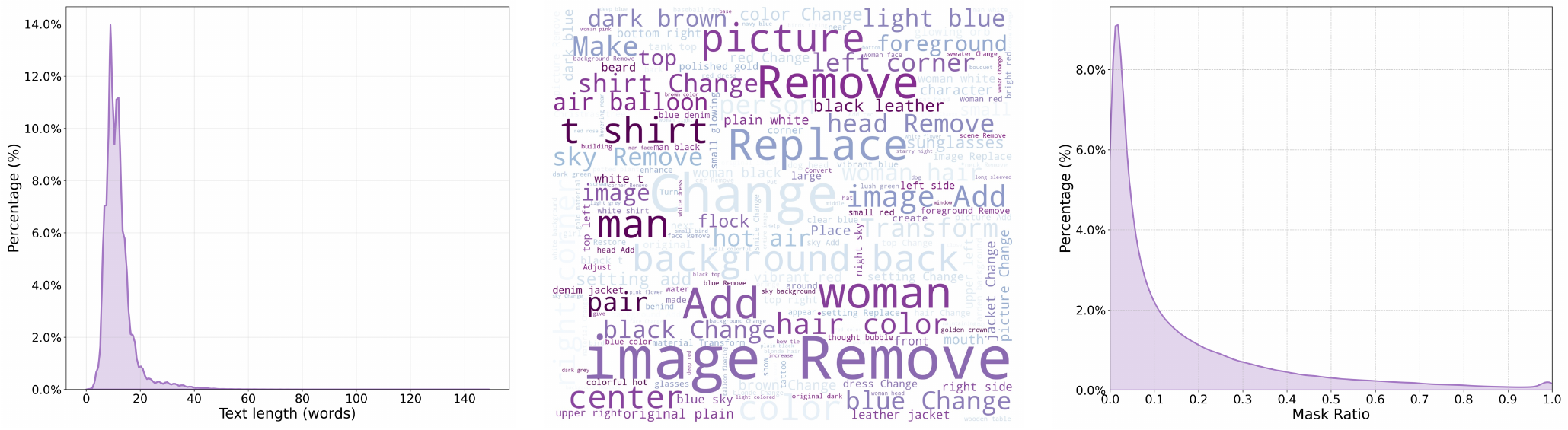}
    \vspace{-1.5em}
    \caption{\textbf{Statistical analysis of GVCoT-Edit-Instruct.} \textbf{Left:} Length distribution of textual editing instructions. \textbf{Middle:} Word cloud of all editing instructions. \textbf{Right:} Proportion of edit-region masks that cover the entire image.}
    \vspace{-1.0em}
    \label{fig:data_statistics}
\end{figure*}

\section{Dataset Details}
\noindent \textbf{Data Curation Pipeline:} During the first stage of our data curation pipeline, we use Qwen2.5-VL-72B \cite{bai2025qwen2} to generate diverse image editing instructions and filter out low-quality triplets, \textit{i.e.}, source image, edit image, and edit instruction. The detailed prompt templates are displayed in Tab. \ref{tab:edit_prompt} and Tab. \ref{tab:eval_sc_prompt}, respectively. In the second stage, we explore the edit region coordinates via Qwen2.5-VL-72B \cite{bai2025qwen2}, the prompt is shown in Tab. \ref{tab:grounding_prompt}.

\begin{figure*}[t]
    \centering
    \includegraphics[width=\linewidth]{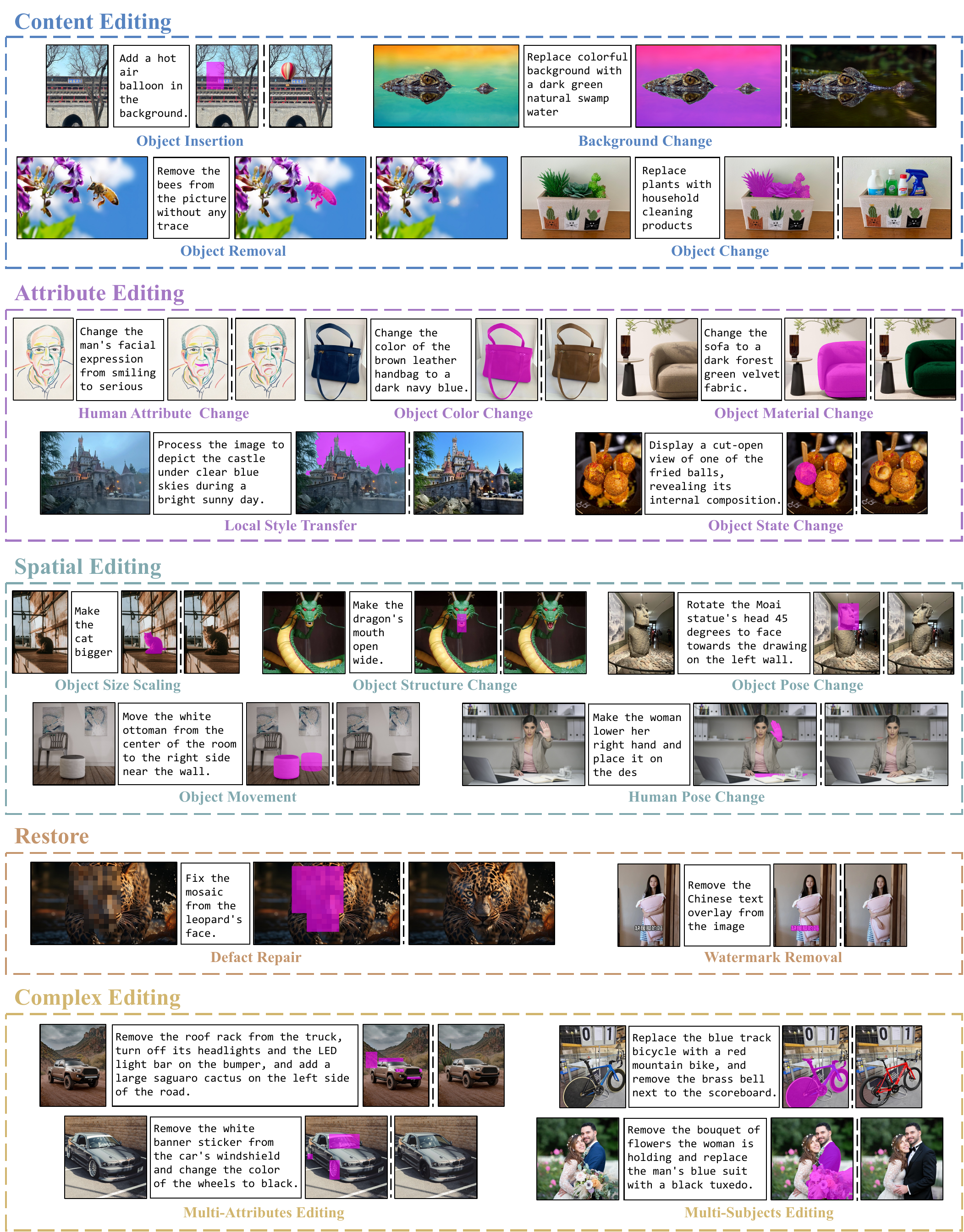}
    \vspace{-1.5em}
    \caption{\textbf{More examples of our training dataset GVCoT-Edit-Instruct.} Each sample consists of four components: (1) the source image, (2) the textual edit instruction, (3) the edit mask, a pink overlay indicating the edit region, and (4) the target image after editing.}
    \label{fig:data_vis}
\end{figure*}

\noindent \textbf{GVCoT-Edit-Instruct:} More statistics are provided in Fig. \ref{fig:data_statistics}, and additional examples are shown in Fig. \ref{fig:data_vis}. Our training set spans a broad spectrum of editing tasks, \textit{i.e.}, content, attribute, spatial, restoration, and complex, covering diverse subjects, such as humans, animals, and things.

\begin{figure*}[t]
    \centering
    \includegraphics[width=\linewidth]{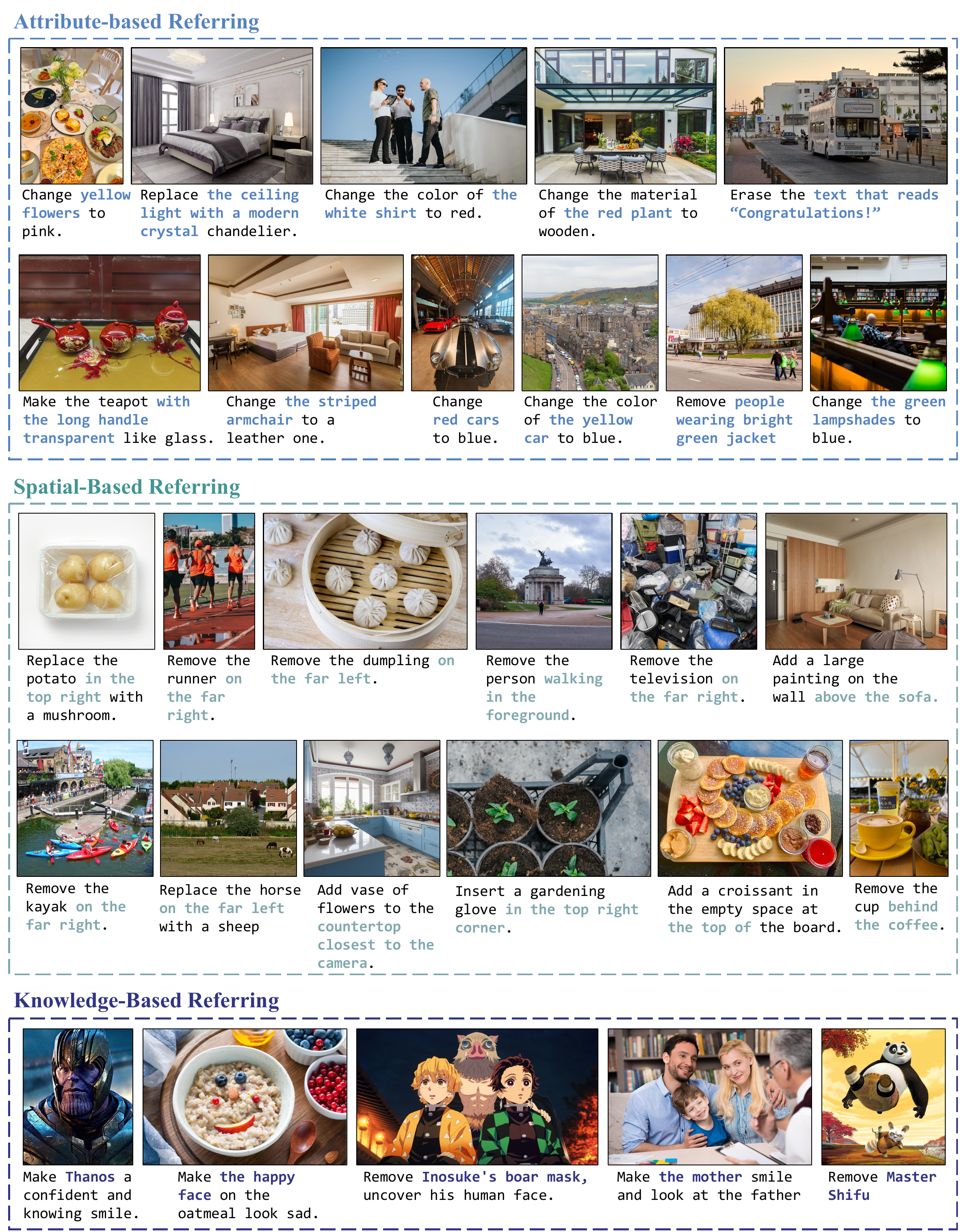}
    \vspace{-1.5em}
    \caption{\textbf{More examples of SREdit-Bench.} We focus on fine-grained referring expressions and sophisticated visual scenes.}
    \label{fig:benchmark_supp}
\end{figure*}

\noindent \textbf{SREdit-Bench:} More examples are shown in Fig. \ref{fig:benchmark_supp}. 

\section{More Experiments}

\begin{figure}
    \centering
    \includegraphics[width=\linewidth]{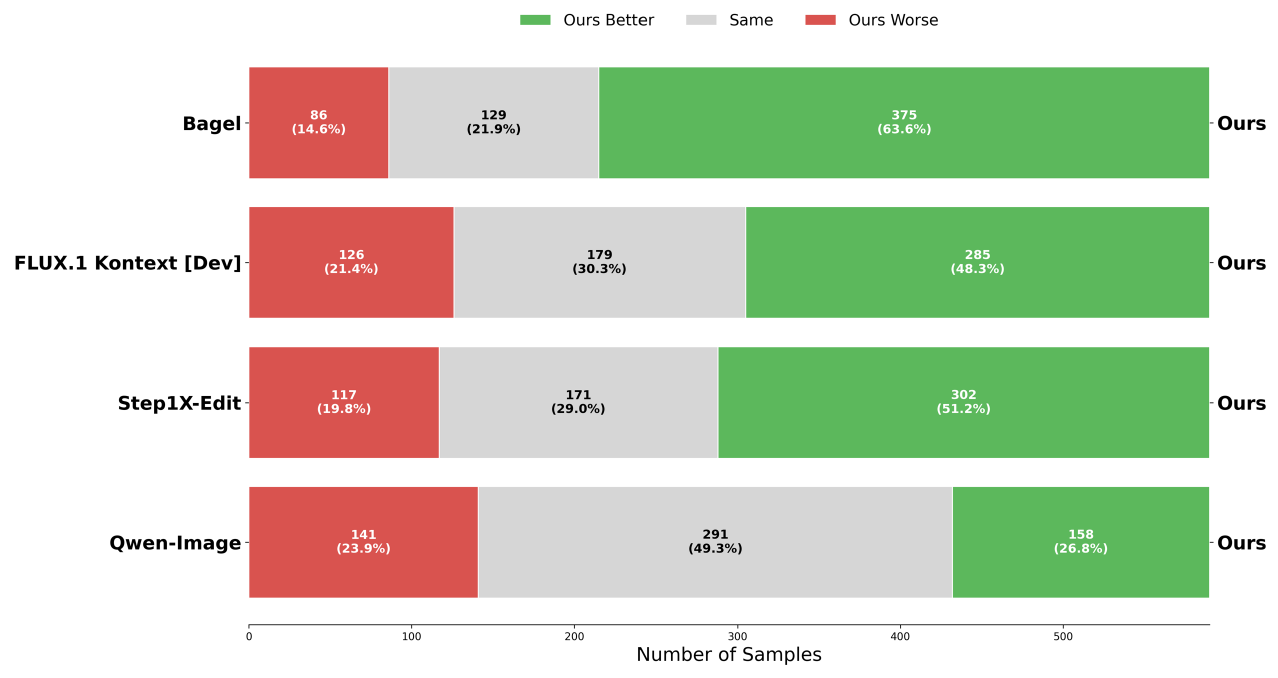}
    \vspace{-1.3em}
    \caption{\textbf{User study.} We conduct GSB (Good/Same/Bad) to compare our method with state-of-the-art models.}
    \vspace{-1.5em}
    \label{fig:user_study}
\end{figure}

\noindent \textbf{User Study.} 
We conduct a comprehensive user study on our SREdit-Bench involving 56 participants with diverse backgrounds, including graduate students and practitioners familiar with image editing. We employed the GSB (Good/Same/Bad) metric to assess our method against four leading baselines: Qwen-Image \cite{wu2025qwenimagetechnicalreport}, Step1X-Edit \cite{liu2025step1x}, and FLUX.1 Kontext [Dev] \cite{labs2025flux1kontextflowmatching}, and Bagel \cite{deng2025bagel}. In this blind A/B testing setup, participants evaluated pairwise results to determine if our generation was superior (Good), comparable (Same), or inferior (Bad) to the competitors. Fig. \ref{fig:user_study} shows that our methods achieve a clear advantage.

\noindent \textbf{Additional results on GEdit-Bench, MagicBrush, and Emu Edit.} 
We also evaluate our Bagel-GVCoT under other benchmarks in Tab. \ref{tab:gedit} and Tab. \ref{tab:magicbrush}. Our Bagel-GVCoT consistently outperforms the baseline across all metrics. These results collectively confirm that our proposed reasoning paradigm enhances the base model's ability to perform high-fidelity and instruction-consistent image editing.

\begin{figure*}[t]
    \centering
    \includegraphics[width=\linewidth]{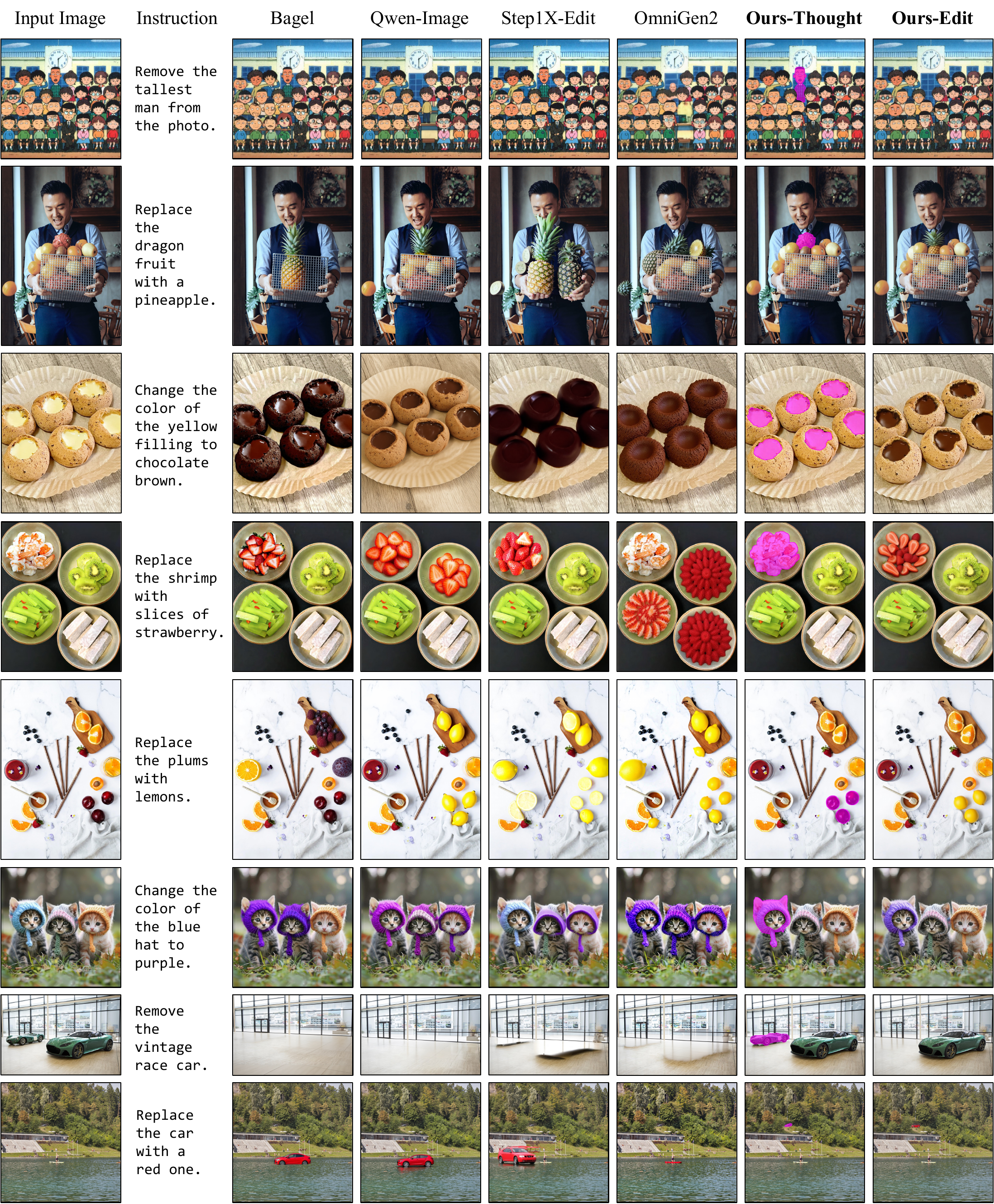}
    \vspace{-1.0em}
    \caption{\textbf{Qualitative comparison in SREdit-Bench.} Our method demonstrates superior spatial reasoning and instruction adherence compared to existing open-source models, especially when handling complex, multi-object editing tasks. }
    \label{fig:results2}
\end{figure*}

\noindent \textbf{Additional qualitative results.} 
More qualitative results in complex scenes that demand strong spatial reasoning are shown in Fig. \ref{fig:results2}. Both Qwen-Image \cite{wu2025qwenimagetechnicalreport} and Step1X-Edit \cite{liu2025step1x} struggle to correctly identify the intended edit region. Our Bagel-GVCoT consistently captures fine-grained spatial cues and produces more faithful region localization than the baseline. This advantage arises from our explicit spatial reasoning pathway, which enables the model to determine where to edit before executing the manipulation.

\begin{table*}[ht]
\centering
\caption{Key hyperparameters for the two-stage training process in our Progressive Supervised Fine-Tuning. ``\texttt{N/A}'' indicates that the parameter did not apply to that stage.}
\vspace{-0.5em}
\resizebox{0.7\linewidth}{!}{
\begin{tabular}{lcc}
\toprule
\textbf{Hyperparameter} & \textbf{Stage 1} & \textbf{Stage 2} \\
\midrule
\multicolumn{3}{l}{\textit{\textbf{Optimizer \& Scheduler}}} \\
Learning Rate (LR) & $2 \times 10^{-5}$ & $1 \times 10^{-5}$ \\
LR Scheduler & Cosine Decay & Cosine Decay \\
Min Learning Rate & $1 \times 10^{-7}$ & $1 \times 10^{-7}$ \\
Warmup Steps & 1,000 & 1,000 \\
Total Training Steps & 80,00 & 12,000 \\
\midrule
\multicolumn{3}{l}{\textit{\textbf{Model \& Loss}}} \\
EMA Decay Rate & 0.000 & 0.995 \\
Rectified-Flow Timestep Shift & 2.0 & 2.0 \\
Cross-Entropy (CE) Loss Weight & \texttt{N/A} & \texttt{N/A} \\
Rectified-Flow (MSE) Loss Weight & 1.0 (Implicit) & 1.0 (Implicit) \\
Frozen Components & Understanding Expert & None \\
\midrule
\multicolumn{3}{l}{\textit{\textbf{Batching \& Tokenization}}} \\
Max Tokens per Batch & 32,768 & 40,000 \\
Max Tokens per Sample & 16,384 & 20,000 \\
\midrule
\multicolumn{3}{l}{\textit{\textbf{Regularization (Dropout)}}} \\
Text Condition Dropout & 0.0 & 0.0 \\
ViT Condition Dropout & 0.0 & 0.0 \\
VAE Condition Dropout & 0.0 & 0.0 \\
\bottomrule
\end{tabular}
}
\label{tab:hyperparams_sft}
\end{table*}

\begin{table*}[ht]
\centering
\caption{Key hyperparameters for the two-stage training process in our Reinforcement-based Refining. ``\texttt{N/A}'' indicates that the parameter did not apply to that stage.}
\resizebox{0.7\linewidth}{!}{
\begin{tabular}{lcc}
\toprule
\textbf{Hyperparameter} & \textbf{Stage 1} & \textbf{Stage 2} \\
\midrule
\multicolumn{3}{l}{\textit{\textbf{Optimizer \& Scheduler}}} \\
Learning Rate (LR) & $1 \times 10^{-6}$ & $1 \times 10^{-6}$ \\
LR Scheduler & Cosine Decay & Cosine Decay \\
Min Learning Rate & $1 \times 10^{-7}$ & $1 \times 10^{-7}$ \\
Warmup Steps & 50 & 50 \\
Total Training Steps & 500 & 450 \\
\midrule
\multicolumn{3}{l}{\textit{\textbf{Reward \& Loss \& Model}}} \\
Rectified-Flow Timestep Shift & 3.0 & 3.0 \\
Reward 1 weight & 0.5 & 0.5 \\
Reward 2 weight & 0.8 & 0.2 \\
Group Size & 24 & 24 \\
KL Weight & 0.001 & 0.001 \\
Rectified-Flow Sampler & SDE & SDE \\
Frozen Components & Understanding Expert & Understanding Expert \\
\midrule
\multicolumn{3}{l}{\textit{\textbf{Batching \& Tokenization}}} \\
Max Tokens per Batch & 32,768 & 32,768 \\
Max Tokens per Sample & 16,384 & 16,384 \\
\bottomrule
\end{tabular}
}
\label{tab:hyperparams_rl}
\end{table*}

\begin{table*}[t]
\centering
\begin{tcolorbox}[colframe=black, colback=gray!5, arc=5mm, boxrule=0.5mm, width=\linewidth]
\begin{tabular}{p{\linewidth}}
You are a professional visual quality evaluator. Your task is to qualitatively assess the spatial consistency between an edited image and a CoT image where a mask is overlaid on the original image (i.e., the mask is drawn/painted onto the original, not a separate binary/probability map). All input images and depicted human figures are synthetic (AI-generated); therefore, privacy and confidentiality are not concerns.
\vspace{0.6em}

\textbf{TASK \& RULES:} \\
Three images and an editing instruction will be provided.
\begin{itemize}
\itemsep0em
\item \texttt{The first image}: Image A, The initial, unmodified AI-generated image.
\item \texttt{The second image}: Image B, the post-edit version of Image A.
\item \texttt{The third image}: Image C, CoT overlay image, a copy of Image A with a visible mask painted on top (e.g., uniform magenta or tinted regions indicating intended edit areas). This overlay is for human-readable guidance only and is not a separable per-pixel mask.
\item  \texttt{Instruction}: The textual instruction describes the editing task.
\end{itemize}
Your primary task is to evaluate how successfully the image has been generated, focusing on visual naturalness and technical artifacts, not content preference or aesthetics. 
\vspace{0.6em}

\textbf{EVALUATION RULES:} \\
Determine where edits actually occurred in Image B relative to Image A (derive an edit region conceptually). Visually compare that edit region to the masked regions in Image C. The assessment is conducted in two perspectives: \textbf{Consistency} and \textbf{Leakage}

\textbf{Consistency:} how well edits align with the masked regions (coverage of intended areas).
\begin{itemize}
    \item 0–2: Edits mostly miss the masked region, or no visible change despite a clear mask.
    \item 3–4: Limited alignment; small patches inside mask, most intended area untouched.
    \item 5–6: Partial alignment; roughly half of the intended area affected, notable gaps.
    \item 7–8: Good alignment; most of the masked area edited with minor omissions.
    \item 9–10: Excellent alignment; edits closely conform to masked regions, near-complete coverage.
\end{itemize}
\textbf{Leakage:} how well edits are contained within masked regions (spill outside mask).
\begin{itemize}
    \item 0–2: Heavy spill; large edits outside mask or global change unrelated to mask.
    \item 3–4: Noticeable spill; multiple regions outside the mask are affected.
    \item 5–6: Some spill; limited areas beyond the mask edited.
    \item 7–8: Minor spill; edits largely contained, slight boundary bleed acceptable.
    \item 9–10: Virtually no spill; edits fully contained within mask.
\end{itemize}
\vspace{0.6em}

\textbf{STEP-BY-STEP GUIDANCE:}
\begin{itemize}
    \item \textbf{Infer the edit region by visually comparing Image A and Image B:} Look for changed colors/tones, added/removed objects, geometry changes, texture updates, or local retouching.
    \item \textbf{Infer the masked intent from Image C:} Identify the colored/tinted overlay area(s) intended for editing. Note mask granularity: coarse blob vs. fine boundaries; whether it covers entire objects or subparts.
    \item \textbf{Estimate overlap:} observe within-mask coverage and outside-mask spill
\end{itemize}
\vspace{0.6em}

\textbf{OUTPUT FORMAT:} \\
Your output must be a single, valid JSON object. Keep your reasoning concise; do not include any extra fields, lists, or commentary outside the JSON.
\begin{center}
\begin{verbatim}
{
    "score" : [...], 
    "reasoning" : "..."
}
\end{verbatim}    
\end{center}
Put the score in a list such that output \texttt{score = [score1, score2]}, where \texttt{score1} evaluates the consistency and \texttt{score2} evaluates the leakage.

\end{tabular}
\end{tcolorbox}
\caption{
The prompt template used by Qwen2.5-VL-72B \cite{bai2025qwen2} to measure the consistency of the edited image with CoT visual thought.}
\label{tab:eval_gvcot_prompt}
\end{table*}

\begin{table*}[t]
\centering
\begin{tcolorbox}[colframe=black, colback=gray!5, arc=5mm, boxrule=0.5mm, width=\linewidth]
\begin{tabular}{p{\linewidth}}
You are a professional digital artist and visual quality evaluator. Your task is to assess an AI-generated image for overall naturalness and visible artifacts. All input images and depicted human figures are synthetic (AI-generated); therefore, privacy and confidentiality are not concerns.
\vspace{0.6em}

\textbf{TASK \& RULES:} \\
Two images and an editing instruction will be provided.
\begin{itemize}
\itemsep0em
\item \texttt{The image}: The initial, unmodified AI-generated image.
\end{itemize}
Your primary task is to evaluate how successfully the image has been generated, focusing on visual naturalness and technical artifacts, not content preference or aesthetics. 
\vspace{0.6em}

\textbf{EVALUATION RULES:} \\
The assessment is conducted in two perspectives: \textbf{Naturalness} and \textbf{Artifacts}.

\textbf{Naturalness (0–10): How realistic or internally consistent the image appears (including non-photoreal styles).}
\begin{itemize}
    \item 0: 0: Extremely unnatural; severe violations of physics, geometry, or lighting.
    \item 1–3: Clearly unnatural; major issues in perspective, scale, light/shadow, or materials.
    \item 4–6: Partially natural; broadly acceptable but with notable inconsistencies (e.g., shadow direction mismatch, implausible depth).
    \item 7–8: Mostly natural; only minor issues (e.g., slight edge artifacts, imperfect reflections).
    \item 9–10: Highly natural; lighting, materials, perspective, and scene relations are coherent with minimal to no discordance.
\end{itemize}
\textbf{Artifacts (0–10): Degree of visible technical flaws (higher is fewer flaws).}
\begin{itemize}
    \item 0–2: Heavy artifacts; strong distortion, watermarks, severe smearing, structural breaks.
    \item 3–4: Many artifacts; multiple blurs, compression noise, repeated textures, broken edges.
    \item 5–6: Moderate artifacts; localized noise, mild ghosting, texture stretching, banding.
    \item 7–8: Minor artifacts; small areas of inconsistency or slight detail loss.
    \item 9–10: Virtually no artifacts; clean image with no clear technical issues.
\end{itemize}
\vspace{0.6em}

\textbf{CHECKLISTS:}
\begin{enumerate}
    \item \textbf{Geometry and perspective:} Are proportions and perspective plausible? Are structures continuous and counts correct (fingers, ears, door frames, wheels)?
    \item \textbf{Color and exposure:} Is white balance, contrast, and dynamic range natural? Any over-sharpening or over-smoothing? Any banding or moiré? 
    \item \textbf{Physics and lighting:} Are light direction, intensity, and color temperature consistent? Do shadows match positions and shapes? Are reflections/refractions plausible?
    \item \textbf{Anatomy and biology:} Finger count and shape, eyes/teeth details, limb connections correct; face free of synthesis artifacts. Compression blocks, noise, grid patterns, repeated motifs, jagged edges, paint spill/bleeding, watermarks, or residual logos.
\end{enumerate}
\vspace{0.6em}

\textbf{OUTPUT FORMAT:} \\
Your output must be a single, valid JSON object. Keep your reasoning concise; do not include any extra fields, lists, or commentary outside the JSON.
\begin{center}
\begin{verbatim}
{
    "score" : [...], 
    "reasoning" : "..."
}
\end{verbatim}    
\end{center}
Put the score in a list such that output \texttt{score = [score1, score2]}, where \texttt{score1} evaluates the naturalness and \texttt{score2} evaluates the degree of artifacts.

\end{tabular}
\end{tcolorbox}
\caption{
The prompt template used by Qwen2.5-VL-72B \cite{bai2025qwen2} to measure the perceptual quality of source and edited images.}
\label{tab:eval_pq_prompt}
\end{table*}

\begin{table*}[t]
\centering
\begin{tcolorbox}[colframe=black, colback=gray!5, arc=5mm, boxrule=0.5mm, width=\linewidth]
\begin{tabular}{p{\linewidth}}
\textbf{You are an expert in image editing and instruction generation.} Given an image, your task is to generate a single, concise, and comprehensive instruction that describes significant editing operations.
\vspace{0.6em}

\textbf{RULES \& OBJECTIVE}
\begin{enumerate}
    \item \textbf{Generate forward instruction:} Based on the content in a provided source image, create a clear, specific, and actionable editing instruction to transform the source image into a hypothetical edited image.
    \item \textbf{Generate reverse instruction:} Create a corresponding reverse editing instruction that transforms the hypothetical image back into the source image.
\end{enumerate}
\vspace{0.6em}

\textbf{GENERAL INSTRUCTION GUIDELINES:}
\begin{enumerate}
    \item \textbf{Conciseness:} Be as brief as possible while ensuring all important edits are included. Always use concise language.
    \item \textbf{Completeness:} If multiple significant edits are present, describe all major changes together in a single, clear, and concise instruction. \textbf{Do not omit any important transformation.}
    \item \textbf{Generality:} If a change can be described with a broad or general term, avoid unnecessary specifics. For example, if a new bank card appears, state: \texttt{Add a bank card without describing the numbers}.
    \item \textbf{Perspective:} Ignore minor changes in text details. Always refer to left or right hands from the \textbf{subject's own perspective}, not the viewer's.
    \item \textbf{Context Independence:} The reversed instruction must not rely on knowledge of the original image or the forward instruction (\textit{e.g.}, do not use phrases like: \texttt{add back the deleted X} \quad \texttt{delete the added Y}).
\end{enumerate}
\vspace{0.6em}

\textbf{MAJOR EDIT CATEGORIES:} \\
The editing instruction must fall into one of the following five precise categories.
\begin{enumerate}
    \item \textbf{Add/Remove Object}
    \begin{itemize}[label=\(\cdot\), itemsep=0em]
        \item \textit{Goal:} Describe the appearance of new objects or the disappearance of existing ones.
        \item \textit{Method:} Specify the object and its location. For partially visible or unidentifiable new objects, use general terms (e.g., letters, a shape, an object) and specify their location. For example: \texttt{Add a dog sitting next to the boy}
        \item \textit{Avoid:} Describing changes in object attributes or guessing the exact identity of partially visible objects.
    \end{itemize}
    \item \textbf{Replace/Transform Object}
    \begin{itemize}[label=\(\cdot\), itemsep=0em]
        \item \textit{Goal:} Describe objects that have been swapped, replaced, or fundamentally transformed.
        \item \textit{Method:} Specify the object replaced/transformed and its new identity. Use the \textbf{subject's perspective} for left/right-hand descriptions. For example, \texttt{Add a dog sitting next to the boy}
    \end{itemize}
    \item \textbf{Color/Style Adjustment}
    \begin{itemize}[label=\(\cdot\), itemsep=0em]
        \item \textit{Goal:} Describe modifications to the color, brightness, saturation, or artistic style of objects in images. Do not describe the addition, removal, or replacement of objects.
        \item \textit{Method:} Specify which object or area had its color or style changed, and describe the new appearance. (\textit{e.g.}, \texttt{Change the car color from red to blue}).
    \end{itemize}
    \item \textbf{Resize/Relocate/Rotate Object}
    \begin{itemize}[label=\(\cdot\), itemsep=0em]
        \item \textit{Goal:} Focus only on whether objects have been resized (enlarged or reduced), moved to a different location, or rotated (clockwise or counterclockwise). Do not describe changes in object identity, color, or style.
        \item \textit{Method:} Specify which object was resized, relocated, or rotated, and describe the new size, position, or orientation. For example, \texttt{Move the tree to the right side of the house.}
    \end{itemize}
    \item \textbf{Change Background}
    \begin{itemize}[label=\(\cdot\), itemsep=0em]
        \item \textit{Goal:} Describe changes to the background or setting of the image. Focus only on modifications to the environment, scenery, or backdrop. Do not describe changes to foreground objects.
        \item \textit{Method:} Specify how the background has changed, such as a new scene, color, or setting. For example, \texttt{Replace the background with a beach scene.}
    \end{itemize}
\end{enumerate}
\vspace{0.6em}

\textbf{OUTPUT FORMAT:}\\
You are required to output the XML content \textbf{directly} in the specified format, without any surrounding text, prefixes, suffixes, or code delimiters.
The output structure must strictly adhere to the following XML template:
    \texttt{<result><type>xxx</type><instruction>xxx</instruction>} \\ \texttt{<reverse>xxx</reverse></result>}

\end{tabular}
\end{tcolorbox}
\vspace{-0.5em}
\caption{
The prompt template used by Qwen2.5-VL-72B \cite{bai2025qwen2} for generating diverse image editing instructions.}
\label{tab:edit_prompt}
\end{table*}

\begin{table*}[t]
\centering
\begin{tcolorbox}[colframe=black, colback=gray!5, arc=5mm, boxrule=0.5mm, width=\linewidth]
\begin{tabular}{p{\linewidth}}
You are a highly skilled digital art quality and integrity evaluator. All input images and depicted human figures are synthetic (AI-generated); therefore, privacy and confidentiality are not concerns.
\vspace{0.6em}

\textbf{TASK \& RULES:} \\
Two images and an editing instruction will be provided.
\begin{itemize}
\itemsep0em
\item \texttt{The first image}: Image A, the initial, unmodified AI-generated image.
\item \texttt{The second image}: Image B, the post-edit version of Image A.
\item \texttt{Instruction}: The textual instruction describes the editing task.
\end{itemize}
Your primary task is to evaluate how effectively an AI-generated image has been edited according to a given instruction. Notice that sometimes Image B may look identical to Image A due to failed editing or subtle changes.
\vspace{0.6em}

\textbf{EVALUATION RULES:} \\
The assessment is conducted in two perspectives: \textbf{Editing success} and \textbf{Over-editing degree}.

\textbf{Editing success scores (0-10): How well Image B fulfills the editing instruction.}
\begin{itemize}
    \item 0: No part of the instruction is followed, or the change contradicts it.
    \item 1–3: Minimal or incorrect attempt; major elements missing/misapplied.
    \item 4–6: Partial success; key aspects present but with notable errors or omissions.
    \item 7–8: Mostly correct; minor inaccuracies or small missing details.
    \item 9–10: Fully correct; aligns precisely with instruction, including details and constraints.
\end{itemize}
\textbf{Overediting degree (0–10): How minimally and appropriately Image B changes non-requested aspects of Image A while still achieving the instruction.}
\begin{itemize}
    \item 0–2: Heavily changed overall; many unintended alterations (composition, subjects, style, lighting) unrelated to the instruction.
    \item 3–4: Major unintended changes, though some original elements remain.
    \item 5–6: Moderate spillover changes; noticeable but not excessive.
    \item 7–8: Light, localized changes; most of the original is preserved.
    \item 9–10: Minimal, surgical edit; only what’s necessary is changed.
\end{itemize}
\vspace{0.6em}

\textbf{STEP-BY-STEP GUIDANCE:}
\begin{enumerate}
    \item \textbf{Parse the instruction:} what exactly must be added/removed/changed (objects, attributes, colors, count, size, position, orientation, lighting, mood, style, text).
    \item \textbf{Compare Image A with Image B:} Did the requested elements appear/disappear/transform as specified? Are spatial relations respected (left/right, in front of, above, near/far, inside/outside)? Were unrelated elements changed (background, palette, composition, subjects) without being required? Did the edit alter the overall scene identity unnecessarily? For small instructions, expect small edits; penalize broad changes. 
    \item \textbf{Edge cases:} If Image B is visually identical to Image A and the instruction required change: editing success score = 0, over-editing degree = 9–10 (minimal but ineffective). If multiple sub-instructions exist, grade holistically but mention critical missing parts.
\end{enumerate}
\vspace{0.6em}

\textbf{OUTPUT FORMAT:} \\
Your output must be a single, valid JSON object. Keep your reasoning concise; do not include any extra fields, lists, or commentary outside the JSON.
\begin{center}
\begin{verbatim}
{
    "score" : [...], 
    "reasoning" : "..."
}
\end{verbatim}    
\end{center}
Put the score in a list such that output \texttt{score = [score1, score2]}, where \texttt{score1} evaluates the editing success and \texttt{score2} evaluates the degree of over-editing.

\end{tabular}
\end{tcolorbox}
\caption{
The prompt template used by Qwen2.5-VL-72B \cite{bai2025qwen2} to measure the semantic consistency of triplets, \textit{i.e.}, source image, edited image, and edit instruction, during our data curation procedure.}
\label{tab:eval_sc_prompt}
\end{table*}

\begin{table*}[t]
\centering
\begin{tcolorbox}[colframe=black, colback=gray!5, arc=5mm, boxrule=0.5mm, width=\linewidth]
\begin{tabular}{p{\linewidth}}
You are a highly skilled digital art forensic analyst. Your core function is to precisely identify and localize editing regions within AI-generated imagery based on comparative analysis. All input images and depicted human figures are synthetic (AI-generated); therefore, privacy and confidentiality are not concerns.
\vspace{0.6em}

\textbf{TASK \& RULES:} \\
Two images and an editing instruction will be provided.
\begin{itemize}
\itemsep0em
\item \texttt{The first image}: The initial, unmodified AI-generated image.
\item \texttt{The second image}: The post-edit version of the original image.
\item \texttt{Instruction}: The textual instruction describes the editing task.
\end{itemize}
Your objective is to locate the specific areas where the edits are conducted, in the \textbf{first original} image. To achieve this, you should carefully analyze the content difference of \textbf{the first original} image and \textbf{the second edited} image, and examine the provided editing instruction text.
\vspace{0.6em}

\textbf{REQUIREMENTS:} \\
You should carefully categorize the editing instruction:
\begin{enumerate}
    \item For insertion of new objects, locate the \textbf{target position} where the object was inserted.
    \item For changes to color, material, and texture, identify and locate the object that has undergone the modification.
    \item  For background replacement, focus on the \textbf{relevant background area} that was changed, not the entire image.
    \item For structure modification, for example: ``Change the person's posture from standing with arms crossed to holding a cup up'', locate the \textbf{object} that has undergone the manipulation.
    \item For global style transfer edits, do nothing.
\end{enumerate}
Be aware that a single editing instruction may require identifying \textbf{multiple distinct regions}.
\vspace{0.6em}

\textbf{OUTPUT FORMATE} \\
You should return one or more bounding boxes. Each bounding box is defined by the coordinates of its top-left and bottom-right corners: \texttt{[x1, y1, x2, y2]}.
The output should be formatted as a Python list of lists, as follows: \texttt{[[xxx, xxx, xxx, xxx]]}. All coordinate values must be populated with numerical content. Empty or non-numeric values are strictly forbidden. The output must not contain any additional prefixes, suffixes, explanatory text, or surrounding characters (e.g., Markdown code blocks, natural language sentences).

\end{tabular}
\end{tcolorbox}
\caption{
The prompt template used by Qwen2.5-VL-72B \cite{bai2025qwen2} for mining bounding box coordinates of the edit region.}
\label{tab:grounding_prompt}
\end{table*}

\clearpage
\onecolumn
\twocolumn
{
    \small
    \bibliographystyle{ieeenat_fullname}
    \bibliography{main}
}